%% file: arxiv.tex
\documentclass[11pt]{article}
\usepackage{arxiv}
\usepackage{times}
\usepackage{url}
\usepackage{latexsym}

\colingfinalcopy 

\usepackage{epsfig}
\usepackage{xspace}
\usepackage{amsmath}
\usepackage{amssymb}

\usepackage{multirow}
\usepackage{enumitem}
\usepackage{algorithm}
\usepackage{booktabs}
\usepackage{subcaption}
\usepackage{color}
\usepackage{colortbl}
\usepackage[dvipsnames]{xcolor}  
\usepackage{array}   
\usepackage{graphics}
\usepackage{arydshln}
\usepackage{calc}

\definecolor{Gray}{gray}{0.93}

\usepackage{comment}

\include{subtex/mlVecMat}

\newcommand{\decoder}{\textsc{\textbf{L}earn from \textbf{E}very\textbf{O}ne}}
\newcommand{\short}{\textsc{LEO}}
\newcommand{\shortstar}{\textsc{LEO+}}

\definecolor{mypink2}{RGB}{219, 48, 122}

\usepackage{titlesec}
\titlespacing{\paragraph}{%
  0pt}{
  0.6\baselineskip}{
  0.6em}

\newlength\myheight
\newlength\mydepth
\settototalheight\myheight{Xygp}
\newcommand*\inlinegraphics[1]{%
  \settototalheight\myheight{Xygp}%
  \settodepth\mydepth{Xygp}%
  \raisebox{-\mydepth}{\includegraphics[height=\myheight]{#1}}%
}

\title{Multi-View Learning for Vision-and-Language Navigation}

\author{Qiaolin Xia\textsuperscript{$\clubsuit$}\thanks{\enskip Authors contributed equally.}\quad Xiujun Li\textsuperscript{$\spadesuit\diamondsuit$}\footnotemark[1]\quad Chunyuan Li\textsuperscript{$\diamondsuit$}\quad Yonatan Bisk\textsuperscript{$\spadesuit\diamondsuit\heartsuit$}\quad\\
\bf{ Zhifang Sui}\textsuperscript{$\clubsuit$}\quad \bf{ Jianfeng Gao}\textsuperscript{$\diamondsuit$}\quad \bf{ Noah A. Smith}\textsuperscript{$\spadesuit\heartsuit$}\quad \bf{ Yejin Choi}\textsuperscript{$\spadesuit\heartsuit$} \\
\textsuperscript{$\spadesuit$}Paul G. Allen School of Computer Science \& Engineering, University of Washington\\
\textsuperscript{$\clubsuit$}MOE Key Laboratory of Computational Linguistics, Peking University\quad\\ \textsuperscript{$\diamondsuit$}Microsoft Research AI\quad \textsuperscript{$\heartsuit$}Allen Institute for Artificial Intelligence\\
{\tt \{xiujun,ybisk,nasmith,yejin\}@cs.washington.edu}\\
{\tt xql@pku.edu.cn\quad \{xiul,chunyl,jfgao\}@microsoft.com}
}

\date{}

\begin{document}

\maketitle

\begin{abstract}
Learning to navigate in a visual environment following natural language instructions is a challenging task because natural language instructions are highly variable, ambiguous, and under-specified. In this paper, we present a novel training paradigm, \decoder{} (\short{}), which leverages multiple instructions (as different views) for the same trajectory to resolve language ambiguity and improve generalization. By sharing parameters across instructions, our approach learns more effectively from limited training data and generalizes better in unseen environments. On the recent Room-to-Room (R2R) benchmark dataset, \short{} achieves 16\% improvement (absolute) over a greedy agent\footnote{Using \textsc{Follower}~\cite{fried2018speaker} as the base agent.} (25.3\% $\rightarrow$ 41.4\%) in {\em Success Rate weighted by Path Length} (SPL). Further, \short{} is complementary to most existing models for vision-and-language navigation, allowing for easy integration with the existing techniques, leading to \shortstar. 
It improves generalization on unseen environments when a single instruction is used in testing, and pushes the R2R benchmark to 62\% when multiple instruction are used.

\end{abstract}

\input{01_intro.tex}

\input{03_method.tex}

\input{04_exp.tex}
\input{05_analysis.tex}
\input{02_related.tex}
\input{07_conclusion.tex}

\bibliographystyle{arxiv}
\bibliography{arxiv}

\newpage
\appendix
\input{08_supp.tex}

\end{document}

%% file: subtex/mlVecMat.tex
\newcommand{\RN}[1]{%
	\textup{\lowercase\expandafter{\it \romannumeral#1}}%
}

\newcommand{\ie}[0]{\emph{i.e., }}

\newcommand{\beq}{\vspace{0mm}\begin{equation}}
\newcommand{\eeq}{\vspace{0mm}\end{equation}}
\newcommand{\beqs}{\vspace{0mm}\begin{eqnarray}}
\newcommand{\eeqs}{\vspace{0mm}\end{eqnarray}}
\newcommand{\barr}{\begin{array}}
\newcommand{\earr}{\end{array}}

\newcommand{\Cmat}{{\bf C}}

\newcommand{\Wmat}[0]{{{\bf W}}}

\newcommand{\av}[0]{{\boldsymbol{a}}}

\newcommand{\cv}[0]{{\boldsymbol{c}}}

\newcommand{\ev}[0]{{\boldsymbol{e}}\xspace}

\newcommand{\hv}[0]{{\boldsymbol{h}}}

\newcommand{\sv}[0]{{\boldsymbol{s}}}

\newcommand{\uv}{\boldsymbol{u}}

\newcommand{\xv}{\boldsymbol{x}}

\newcommand{\zv}{\boldsymbol{z}}

\newcommand{\thetav}{\boldsymbol{\theta}}

\newcommand{\tauv}[0]{{\boldsymbol{\tau}}\xspace}

\newcommand{\E}{\mathbb{E}}

\newcommand{\Xcal}{\mathcal{X}}

\newcommand{\Lcal}{\mathcal{L}}

\newcommand{\Acal}{\mathcal{A}}

\newcommand{\Dcal}{\mathcal{D}}

\newcommand{\Mcal}{\mathcal{M}}

\newcommand{\Scal}{\mathcal{S}}



%% file: 01_intro.tex
\section{Introduction}
\label{sec:intro}

\begin{figure*}[t!]
\small
    \centering
	\begin{tabular}{p{2.5cm}p{2.5cm}p{2.5cm}p{2.5cm}p{2.5cm}p{2.5cm}}
\multicolumn{3}{p{7.2cm}}{
$\mathtt{Instruction~{\bf A}}$: \textit{Walk to the left of the clock and \textcolor{Plum}{down the hallway} to the right. \textcolor{Plum}{Turn right} before the shelf and stop in the doorway of the \textcolor{Plum}{bedroom}}. 
}
&
\multicolumn{3}{@{\hspace{-4mm}}p{7.2cm}}{
$\mathtt{Instruction~{\bf B}}$: \textit{Exit the room going straight. \underline{\textcolor{Plum}{Turn right} go \textcolor{Plum}{down the hallway}} until you get to a black bookcase. \textcolor{Plum}{Turn right} and continue going \textcolor{Plum}{down the hallway} until you get to a \textcolor{Plum}{bedroom}. Wait at the entrance.}}	
\vspace{-3mm}
\\
	    \hspace{-0mm}
        \includegraphics[trim={5cm 5cm 8cm 0cm},clip,angle =-90,scale=0.17]{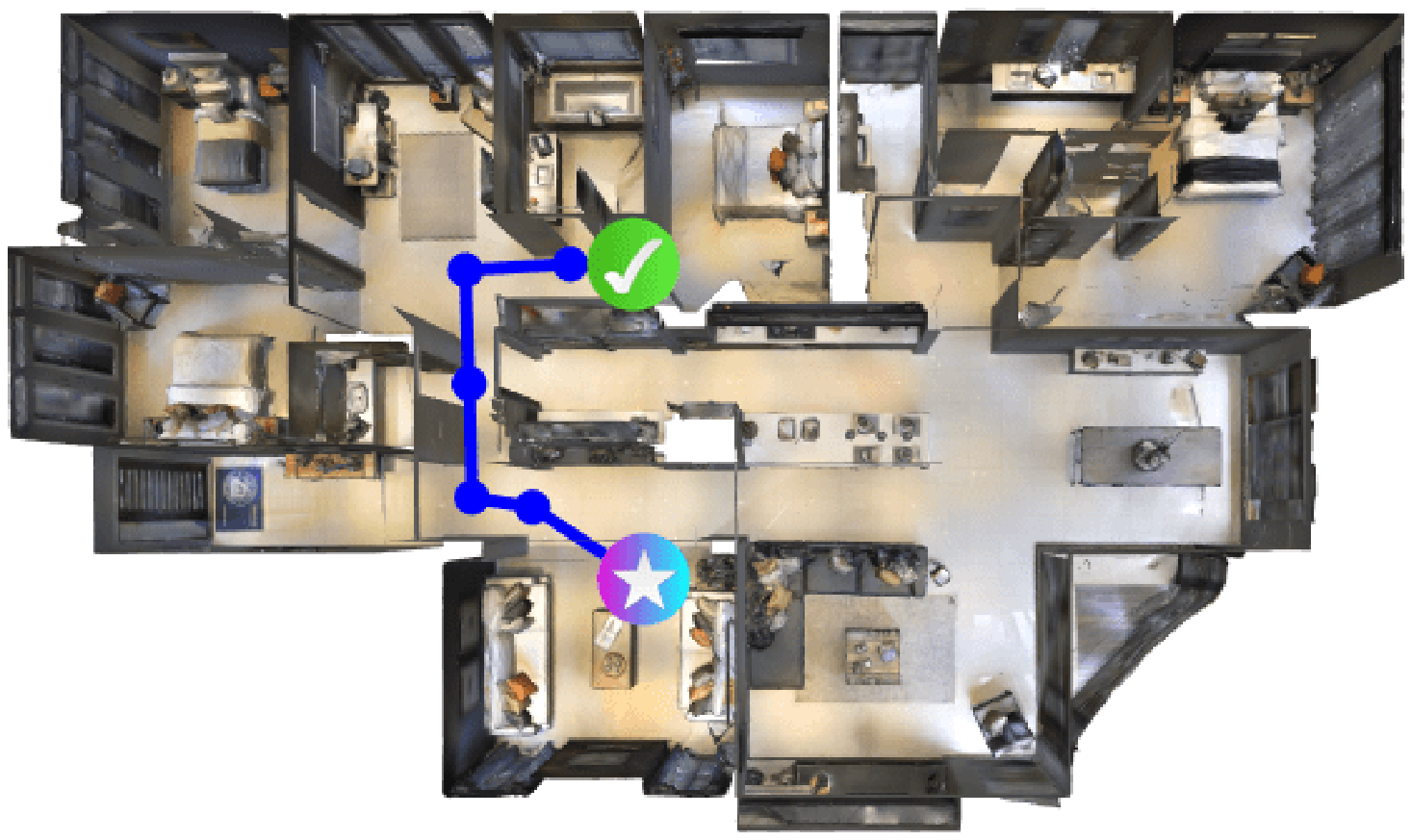} & 
		\hspace{-2mm}
        \includegraphics[trim={5cm 5cm 8cm 0cm},clip,angle =-90,scale=0.17]{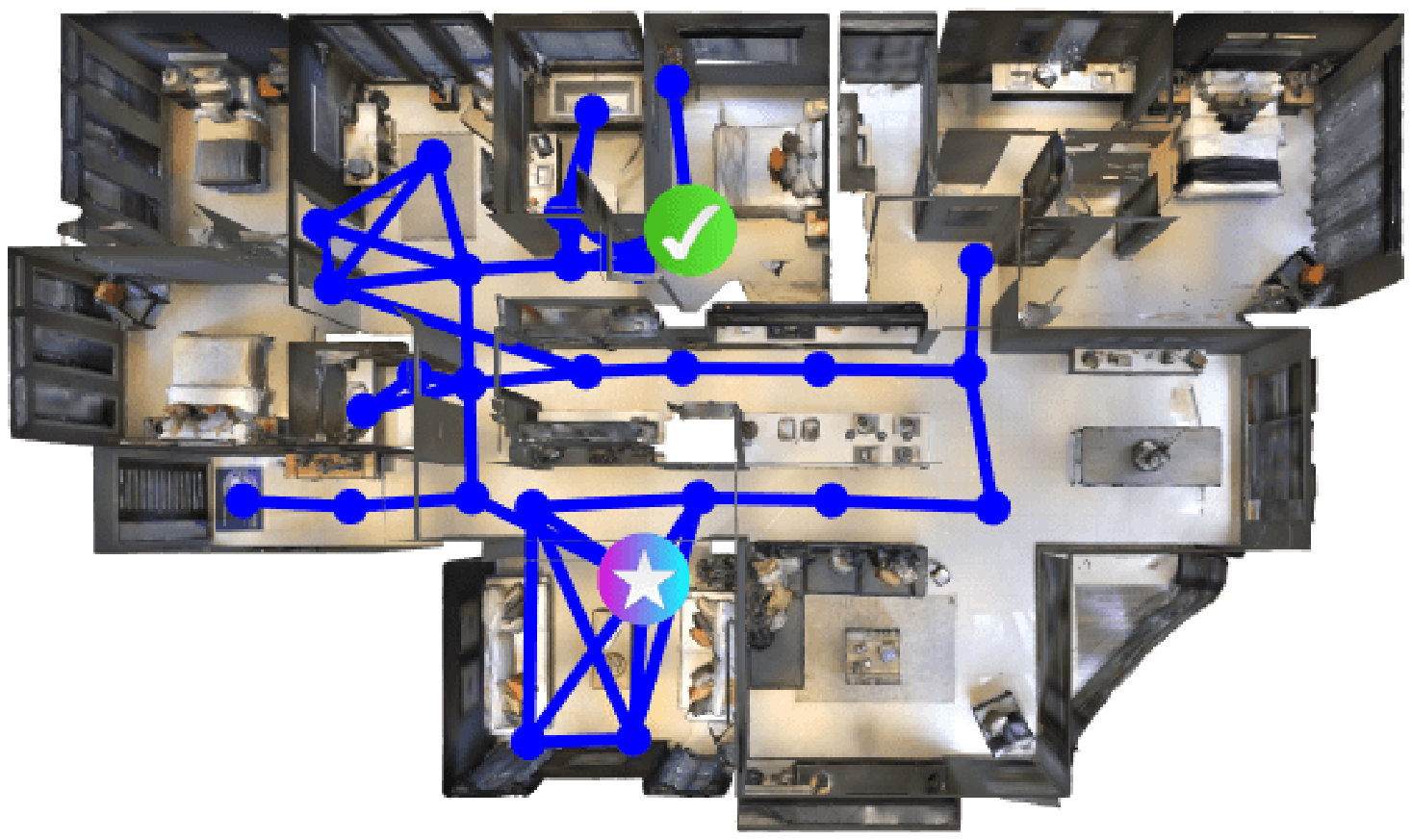} &
		\hspace{-4mm}
        \includegraphics[trim={5cm 5cm 8cm 0cm},clip,angle =-90,scale=0.17]{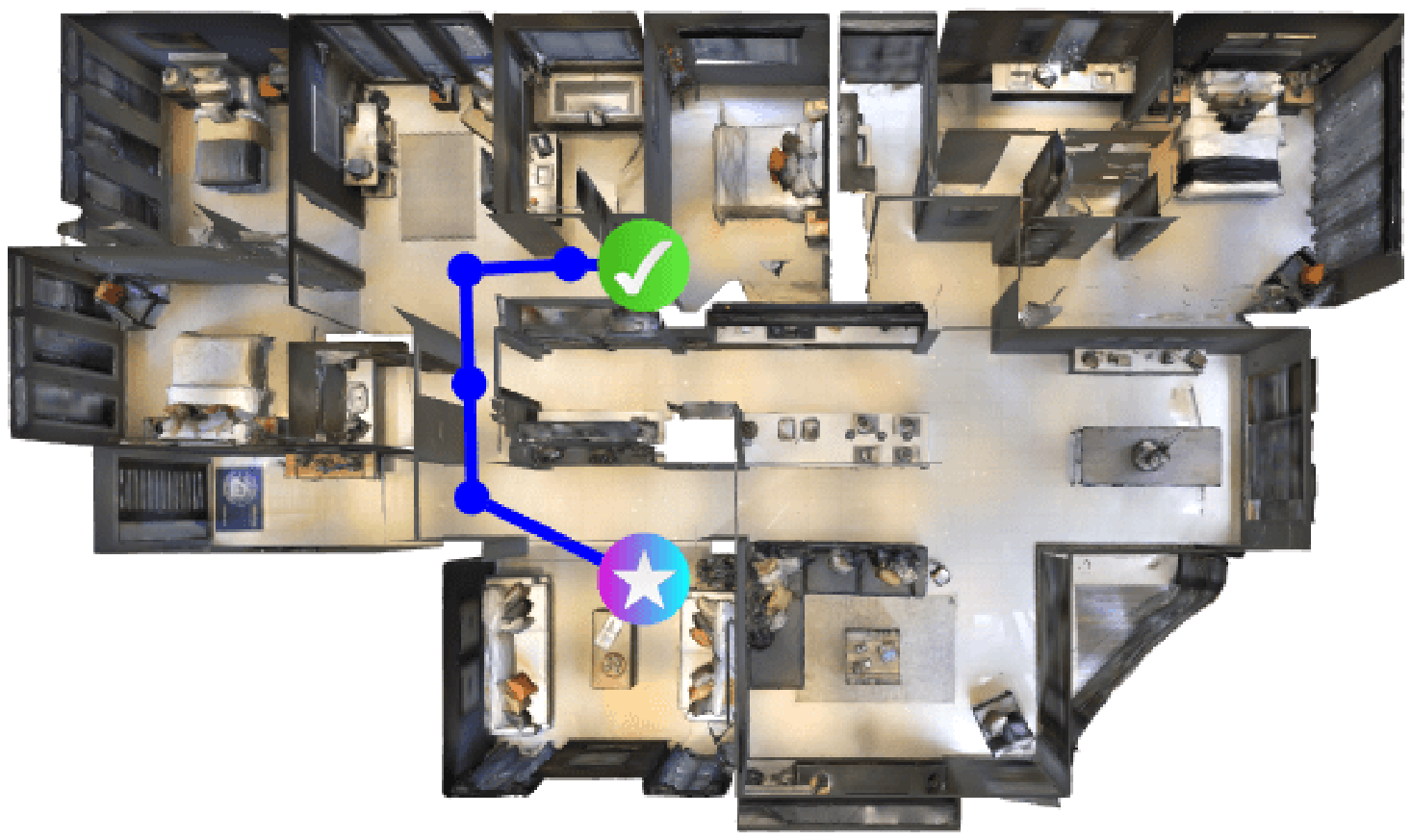} \vspace{-0mm} &
		\hspace{-6mm}
        \includegraphics[trim={5cm 5cm 8cm 0cm},clip,angle =-90,scale=0.17]{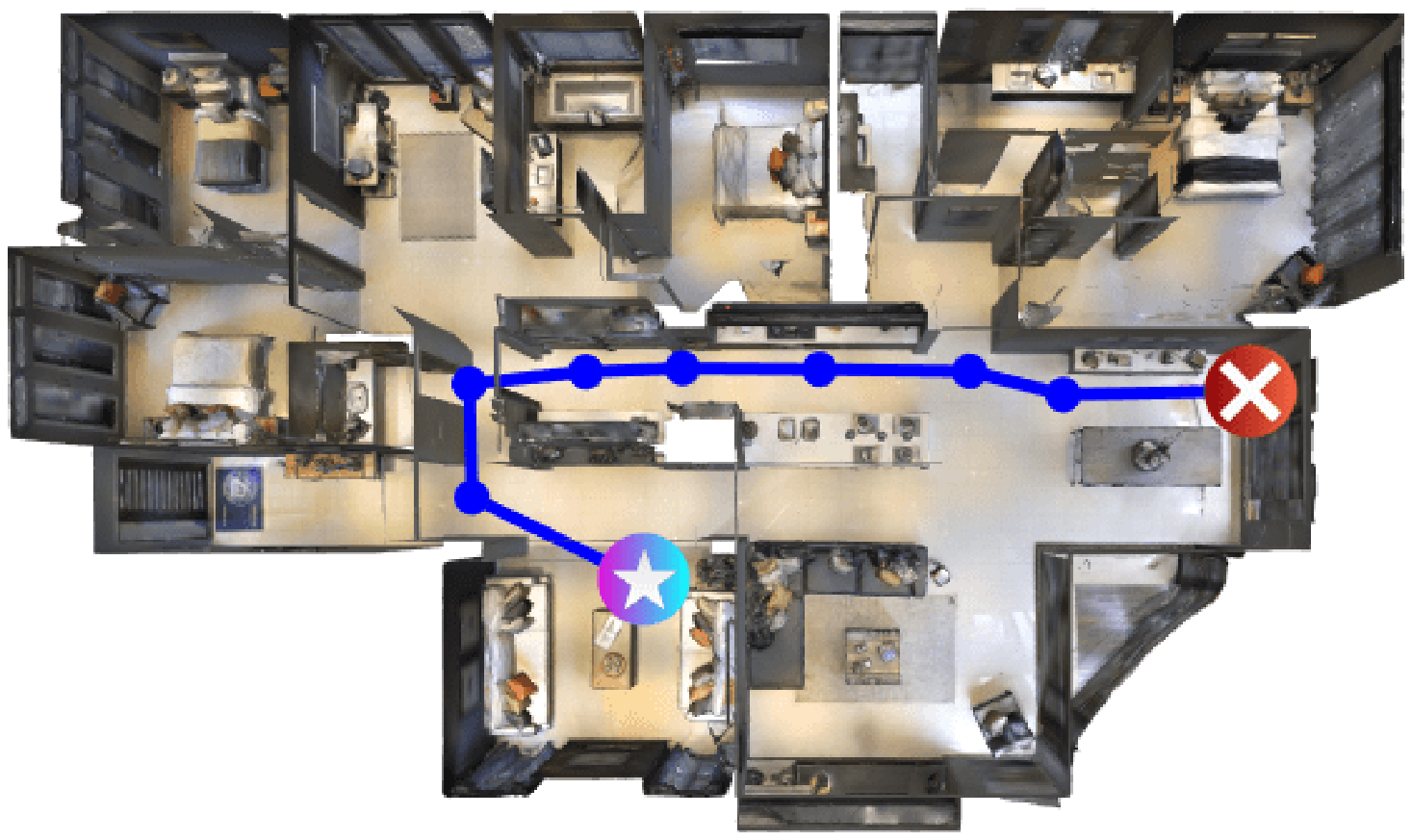}& 
		\hspace{-8mm}
        \includegraphics[trim={5cm 5cm 8cm 0cm},clip,angle =-90,scale=0.17]{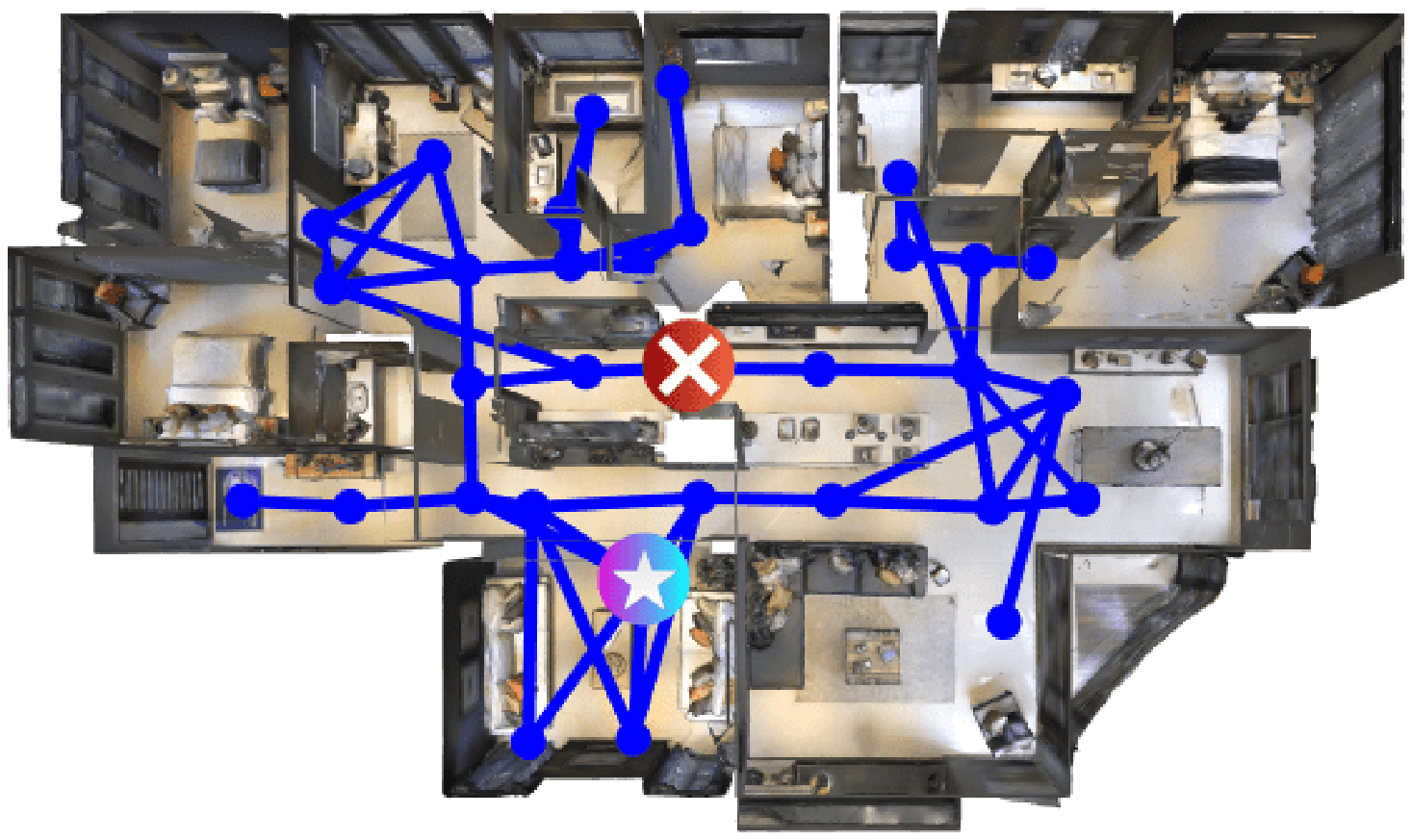} &
		\hspace{-10mm}
        \includegraphics[trim={5cm 5cm 8cm 0cm},clip,angle =-90,scale=0.17]{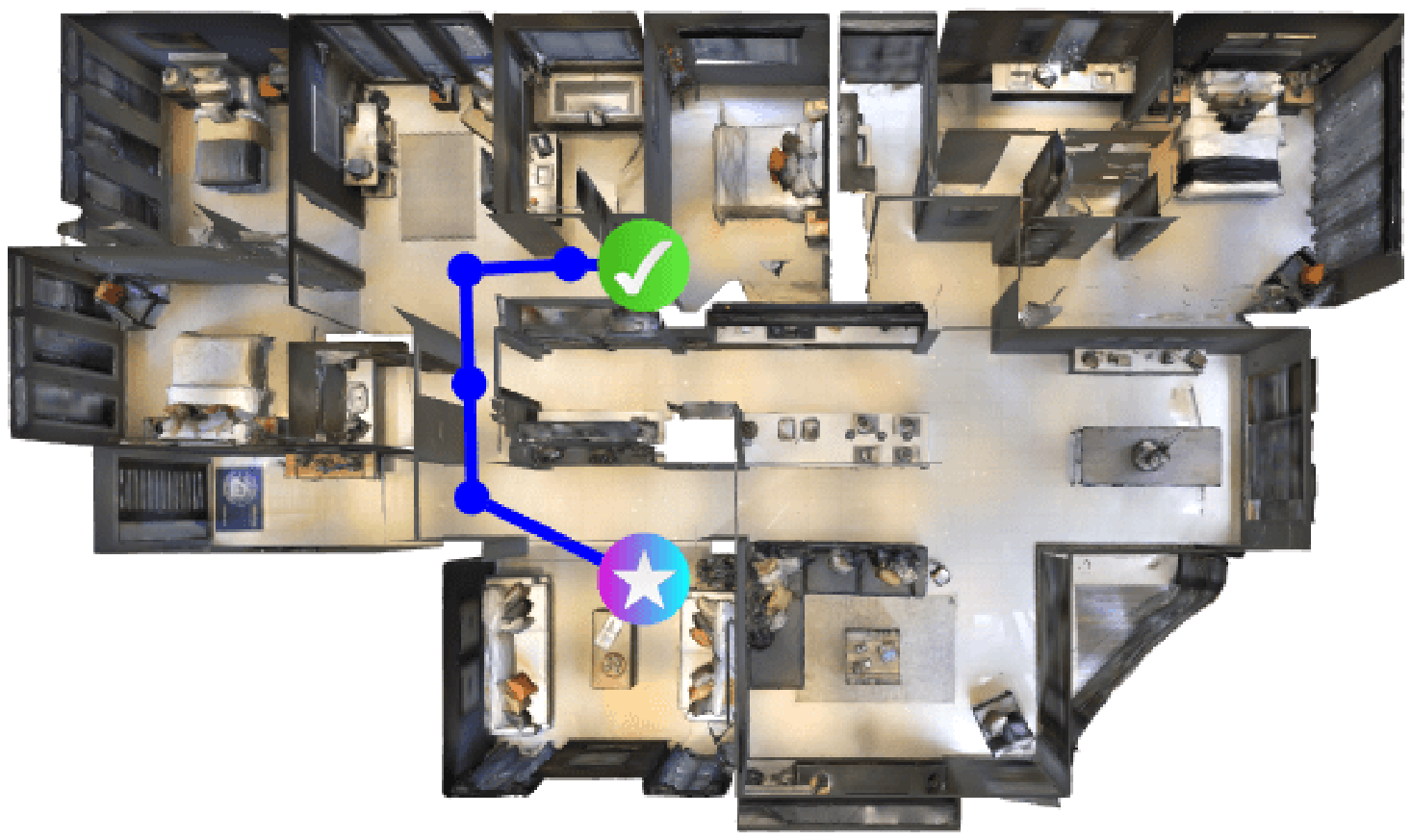}
        \vspace{-0mm} \\
		(a) Greedy agent &
		\hspace{-2mm}
		(b) Beam search &
		\hspace{-4mm}
		(c) Ground-Truth &
		\hspace{-6mm}
		(d) Greedy agent &
		\hspace{-8mm}
		(e) Beam search &
		\hspace{-10mm}
		(f) \short{}		
	\end{tabular}
    \caption{Instructions \textbf{A} and \textbf{B} correspond to the same expert trajectory (c), but use different visual cues. $\mathtt{Instruction~{\bf A}}$ alone is specific enough for agents to reach the target successfully with either a (a) greedy or (b) beam search strategy.
    In contrast, $\mathtt{Instruction~{\bf B}}$, in particular the \underline{underlined} phrase, is ambiguous and causes existing agents (\textit{e}.\textit{g}. \textsc{Speaker-Follower} with beam size=40 (e)) to fail. 
    Our \short{} approach (f) shares information across instructions to learn groundings from ambiguous contexts. We indicate the start 
    (\protect\inlinegraphics{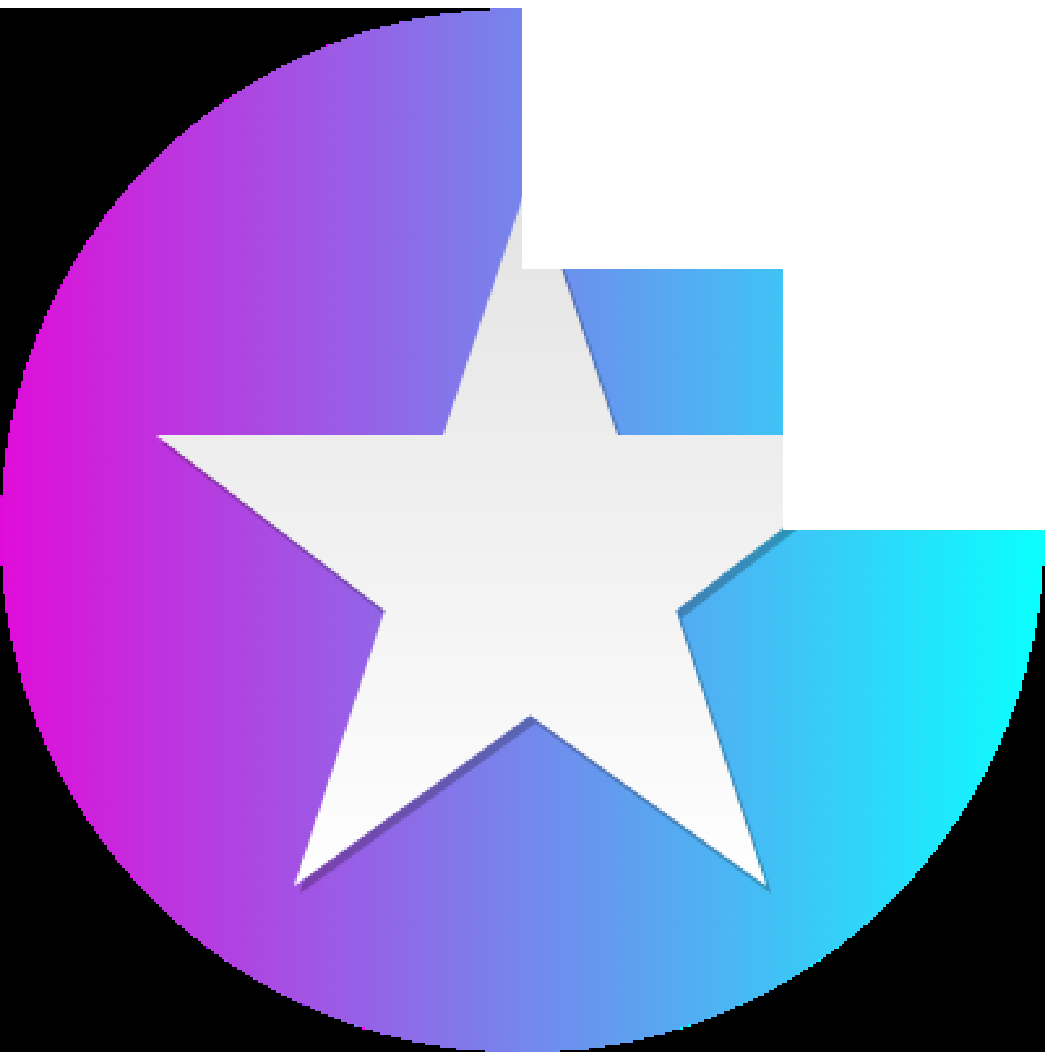}), target 
    (\protect\inlinegraphics{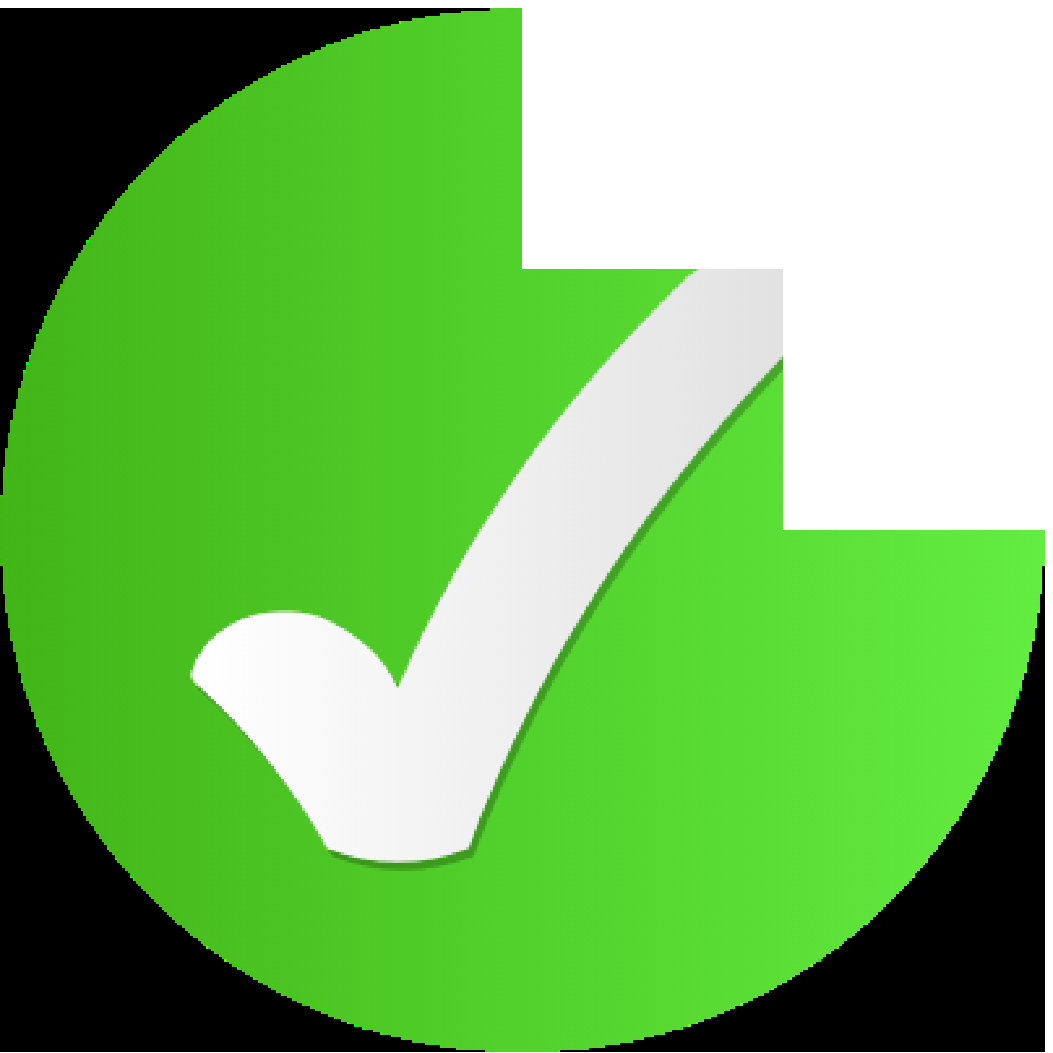}) and failure 
    (\protect\inlinegraphics{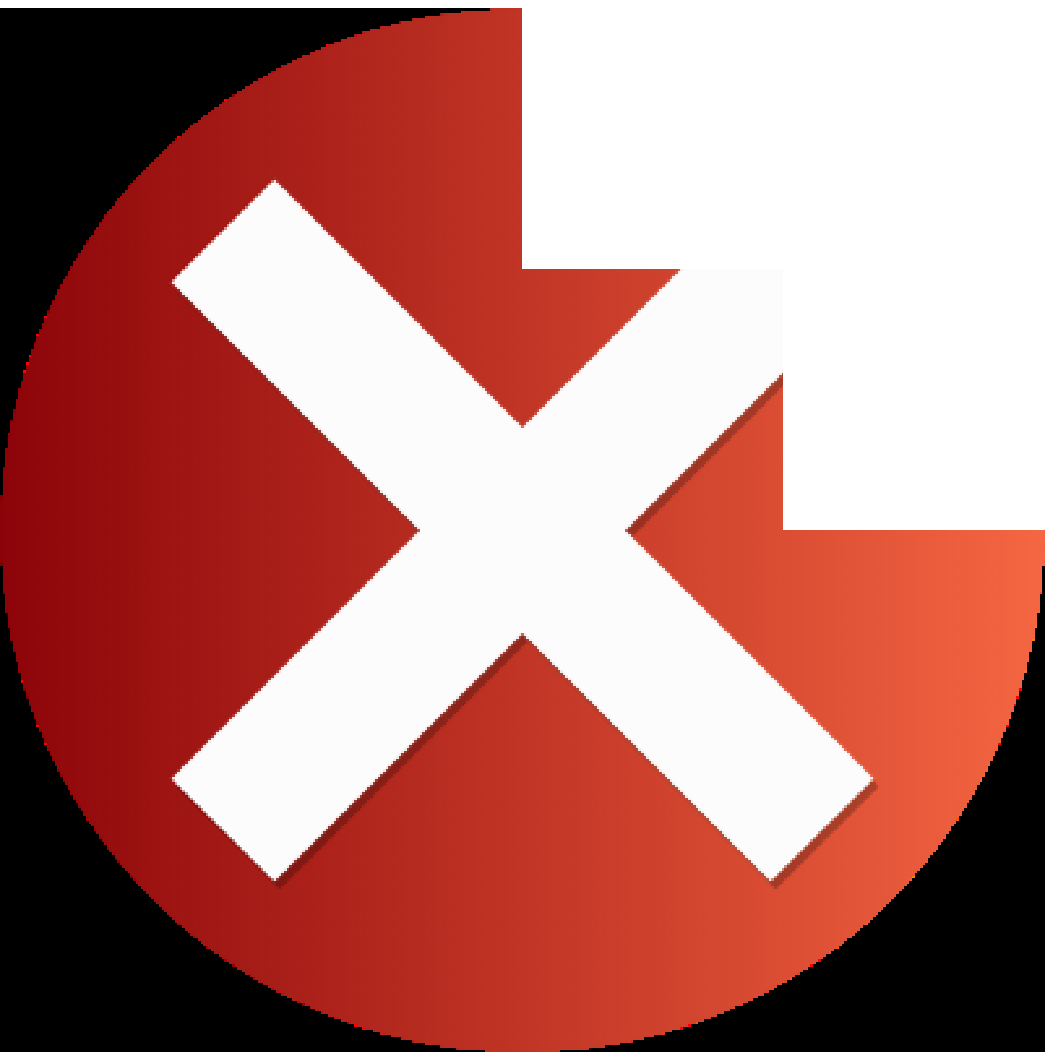}) of agents in an unseen environment.}
    \label{fig:motivation}
\end{figure*}

Learning to navigate in a visual environment based on natural language instructions has attracted increasing research interest in artificial intelligence~\cite{savva2017minos,kolve2017ai2,das2018embodied,anderson2018vision,chen2019touchdown}, as it provides insight into core scientific questions about multimodal representations and takes a step toward real-world applications such as personal assistants and in-home robots.
Navigating from language instructions presents a challenging reasoning problem for agents, as natural language instructions are highly variable, inherently ambiguous, and frequently under-specified. We see this clearly when analyzing the instructions provided in the Room-to-Room (R2R) Vision-and-Language Navigation (VLN) task \cite{anderson2018vision}, where each desired navigation path is paired with highly variable instructions ($\mathtt{Instruction~{\bf A}}$ and $\mathtt{{\bf B}}$ in Figure~\ref{fig:motivation}).


Most previous approaches build on the sequence-to-sequence  architecture~\cite{sutskever2014sequence}, where the instruction is encoded as a sequence of words, and the navigation trajectory is decoded as a sequence of actions, enhanced with better attention mechanisms~\cite{anderson2018vision,wang2018reinforced,ma2019self} and beam search~\cite{fried2018speaker}. 
While a number of approaches~\cite{misra2017mapping,monroe2017colors,wang2018look} have been proposed to reduce the language ambiguity, common to all existing work is that the agent considers each instruction in isolation without collectively reasoning about other alternative instructions for each desired navigation trajectory

However, each instruction in practice only loosely aligns with the desired navigation path, making the existing learning paradigm that considers one instruction at a time less than ideal. This is because every instruction views different components of the trajectory as necessary to mention or obvious to omit, thus they mention and elide different details. For example, sentences like \textit{``Turn right go down the hallway"} ($\mathtt{Instruction~{\bf B}}$ of Figure~\ref{fig:motivation}) may confuse an agent as it is applicable in a wide range of visual contexts (potentially multiple within the same house). 

In order to address this natural variability of instructions more effectively, we propose \decoder{} (\short{}), where each instruction is considered as a different ``view'' of the same ``label'' (trajectory), thus formulating the VLN problem in the ``multi-view'' learning paradigm. The complexity of VLN learning can then be reduced by eliminating hypotheses which lack consensus from multiple views. 
Our new framework, \short{}, effectively aggregates multiple instructions for each navigation trajectory during training. Each instruction plays the role of a teacher, contributing a different clue for completing the task. \short{} allows the agent to fuse the complementary information from multiple teachers when reasoning about the desired trajectory. More concretely, our \short{} navigator encodes multiple instructions of the same trajectory via shared parameterization, and aggregates them via a parameter-free pooling function, before finally generating the action sequence. 

Comprehensive experiments demonstrate strong empirical performance of \short{}, even when we carefully control the number of instructions during testing for fair comparisons. 
On the R2R benchmark dataset,
our new training strategy dramatically improve a seq2seq greedy agent from 25\% to 41\% on SPL in the unseen environments. 
Moreover, since \short{} is easily applicable to most existing approaches to VLN, we also apply \short{} to two state-of-the-art agents: SMNA~\cite{ma2019self} and EnvDrop~\cite{tan2019learning}, obtaining significant improvements: 9\% and 12\% respectively. 
Finally, using the same data augmentation as the existing work, the resulting \shortstar{} navigator achieves 
62\% on the SPL metric,\footnote{Among \emph{all} the public results at the time of this submission.} with 9\% absolute gain over the previous best result.

%% file: 03_method.tex
\section{Preliminaries}
%
The VLN task can be formulated as a Markov Decision Process (MDP) $\Mcal = \left \langle \Scal, \Acal, P_s, r \right \rangle$, where $\Scal$ is the visual state space, $\Acal$ is a discrete action space, $P_s$ is the unknown environment distribution from which we draw the next state, and $r \in \mathbb{R}$ is the reward function. At each time step $t$, the agent first observes an RGB image $\sv_t \in \Scal$, then takes an action $\av_t \in \Acal$. This leads the simulator to generate a new image observation $\sv_{t+1} \sim P_s(\cdot|\sv_t,\av_t)$ as the next state. 
The agent interacts with the environment sequentially, and generates a trajectory of length $T$, $\tauv = [\sv_0, \av_0, \sv_1, \av_1, \cdots, \sv_T, \av_T ] $. 
The episode ends when the agent selects the special $\mathtt{STOP}$ action, or when a pre-defined maximum trajectory length is reached. The navigation is successfully completed, if the trajectory $\tauv$ terminates at the intended target location. 

The success of VLN largely depends on correct grounding of natural language instructions \cite{Thomason:19}. In a typical VLN setting, the instructions are represented as a set $\Xcal = \{\xv_i\}_{i=1}^M$, where $M$ is the number of alternative instructions, and each instruction $\xv_i$ consists of a sequence of $L_i$ word tokens, $\xv_i = [x_{i,1}, x_{i,2}, ..., x_{i,L_i}]$. 
The training dataset $\Dcal_E = \{\tauv, \Xcal\}$ consists of pairs of the instruction set $\Xcal$ together with its corresponding expert trajectory $\tauv$.
%
To simplify the learning, it is widely assumed that the language instructions are {\em independent and identically distributed (iid)}, and therefore each of them can fully represent the task goal. The agent then learns to navigate via performing maximum likelihood estimation (MLE) of the policy $\pi$, based on the individual sequences: 
\begin{align} 
& \max_{\thetav}~ \Lcal_{\thetav}(\tauv, \Xcal) , ~~\mbox{where} \nonumber \Lcal_{\thetav}= \log \pi_{\thetav}(\tauv|\Xcal)
\approx
\frac{1}{M}\sum_{i=1}^M \log \pi_{\thetav}(\tauv|\xv_i),
\label{eq_iid}
\end{align}
and $\thetav$ are the policy parameters. The 
baseline Seq2Seq~\cite{anderson2018vision,fried2018speaker} methods belong to this single teacher learning paradigm.


In this paper, we re-examine the iid assumption in VLN, and raise the concern that policy learning with individual language sequences over-simplifies the dependency 
across alternative instructions, 
leading to degraded performance. 
%

\section{\short{}: Learning from EveryOne}
\label{sec:method}
\subsection{An information theoretic perspective}
We propose to consider the VLN task from the ``multi-view'' learning perspective~\cite{blum1998combining,xu2013survey}, where different instruction contains complementary information to reach the task goal. 
The basic intuition is that the complexity of the learning problem can be reduced considerably by eliminating hypotheses from each instruction that conflict with those of other alternative instructions~\cite{sridharan2008information}.
Thus, we propose \short{}, which learns to navigate based on multiple alternative instructions $\Xcal = \{\xv_i\}_{i=1}^M$ jointly, without making the iid assumption in~\eqref{eq_iid}. 

From an information-theoretic perspective, we can show the connections between the multi-instruction in \short{} and the single instruction training in terms of the conditional entropy (CE):
\begin{align}
\underbrace{ H(\tauv | \Xcal)  }_{~\text{Multi-inst. CE} } 
& = 
\underbrace{ H(\tauv | \xv_1) }_{~\text{Single inst. CE} } 
- \hspace{2mm} 
\underbrace{  I (\tauv, \xv_2, \cdots, \xv_M  | \xv_1 ) }_{~\text{Mutual~Information}}
\end{align}
A detailed proof is provided in Appendix~\ref{sec:proof_of_entr}. It is simple to show that the additional conditioning in \short{} can reduce the entropy in trajectory generation compared to the single instruction setting: 
\begin{align}
H(\tauv | \Xcal) \le H(\tauv | \xv_i), \forall i=1,\cdots, M 
\end{align}
Equality holds if the agent trajectory and other instructions are independent, after the agent observes one specific instruction. We hypothesize this is often false, as many single instructions are too ambiguous to uniquely determine the corresponding agent trajectory. 
\short{} provides the agent an easier task than the single instruction learning, as the prediction uncertainty is reduced when conditioning on more instructions. 


\subsection{\short{} Navigator}
Based on this multi-view learning perspective, we design \short{} navigator on top of the seq2seq baseline~\cite{anderson2018vision,fried2018speaker}. To generate a trajectory, the agent takes action $\av_t$ conditioned on the current state $\sv_t$ and the instruction set $\Xcal$:
%
\begin{align}
\log \pi_{\thetav}(\tauv | \Xcal) = \sum_{t=1}^{T} 
\E_{ P_s( \sv_{t} | \sv_{t-1} , \av_{t-1} )} \log \pi_{\thetav}(\av_t | \sv_t, \Xcal)
= \sum_{t=1}^{T}  \log \pi_{\thetav}(\av_t | \sv_t, \av_{t-1}  \Xcal)
\label{eq_factor}
\end{align}
The second equality holds because the simulator has a deterministic transition function, and we follow~\cite{fried2018speaker} to formulate $\av_{t-1}$ into the policy learning.
Each conditional probability in \eqref{eq_factor} is modeled as a function 
$\pi_{\thetav}(\av_t | \sv_t, \av_{t-1}, \Xcal) = f_{\thetav}( \sv_t, \av_{t-1}, \Xcal) $. In this paper, we use the LSTM~\cite{hochreiter1997long} encoder-decoder framework to parameterize $f_{\thetav}$. The encoder takes all language instructions as input and provides the joint instruction representations. The LSTM decoder predicts the probability over the navigable directions, based on its understanding of the task instruction and the agent's current status.

\paragraph{Trajectory History Context.}
The agent maintains a {\it memory vector} $\hv_t$, summarizing the history of its trajectory $\tauv_{\le t} = [\sv_0, \av_0, \cdots,  \sv_t, \av_t]$ through step $t$.
Once the agent takes a step, the surrounding visual scene changes accordingly.
It first performs one-hop visual attention to look at all of the surrounding view angles, based on its previous memory vector $\hv_{t-1}$. 
Specifically, the current visual state $\sv_t$ is updated as the weighted sum of the panoramic features, $\sv_t = \sum_j \gamma_{t,j} \sv_{t,j}$. The attention weight $\gamma_{t,j}$ for the $j$-th visual feature $\sv_{t,j}$ represents its importance with respect to the previous history context $\hv_{t-1}$, computed as 
$ \gamma_{t,j}= \mbox{Softmax} ( (\Wmat_h \hv_{t-1}  )^{\top} \Wmat_s \sv_{t, j})$~\cite{fried2018speaker} 
where 
$\mbox{Softmax} ( r_{j} ) = \exp( r_{j})/\sum_{j'} \exp( r_{j'})$, $\Wmat_h$ and $\Wmat_s$ are trainable projection matrices. 
Based on $\sv_t$, the agent updates its memory vector via an LSTM:
\vspace{-3mm}
\begin{align}
\hv_t  = f_{\thetav_D}([\sv_t, \av_{t-1}], \hv_{t-1})
\label{eq_memory}
\end{align}
where $\av_{t-1}$ is the action taken at previous step, and $\thetav_{D}$ are the LSTM decoder parameters.

\paragraph{Memory-attended Language Context.} 
Knowing where to navigate requires a dynamic understanding of the language instruction, according to the agent's current status. The memory vector $\hv_t$ enables the agent to keep track of its status, and adaptively focus on whichever part of the instructions are currently most relevant. 

We use an LSTM to encode each language instruction $\xv = [x_1, \cdots, x_L]$ into a sequence of textual features $[\ev_1, \cdots, \ev_L ]$; $\ev_l$ corresponds to the hidden units for word token $x_l$: $\ev_l  = f_{\thetav_E}(x_l, \ev_{l-1})$,
where $\thetav_E$ are the parameters of the LSTM language encoder.

At each time step $t$, the {\it textual context} for the $i$-th instruction $\xv_i$ is computed as weighted sum of word features in the sequence:
\vspace{-3mm}
\begin{align}
\cv_{t,i} = \sum_{l=1}^L \alpha_l \ev_l, ~~\text{where}~~ \alpha_l=\mbox{Softmax}(\hv_t^{\top} \ev_l )
\label{eq_instruction}
\end{align}
Note that $\cv_{t,i}$ places more weight on the words that are most relevant to the agent's current status.

\paragraph{Aggregated Instruction Context.} 
Key to our approach is utilizing the multiple natural language instructions provided for each trajectory as every annotator makes different choices about descriptions/landmarks to include or elide. 
%
At time step $t$, the textual context $\cv_{t,i}$ for different language sequences $\xv_i$ may characterize different aspects of the task instructions. A natural question is how to aggregate $\mathcal{C}_t = \{\cv_{i,t} \}_{i=1}^M$ into a joint context $\zv_t$. 
In this paper, we investigate several parameter-free aggregation functions:
\vspace{-3mm}
\begin{align}
\zv_t = g (\mathcal{C}_t), 
\label{eq_aggregator}
\end{align}
where three schemes are considered:
\begin{itemize}
    \item {\em Mean-pooling}: $\zv_t = \frac{1}{M} \sum_{i=1}^M\cv_{i,t}$; 
    \vspace{-2mm}
    \item {\em Max-pooling}: The most salient features are kept by taking the maximum value along each dimension of the language contexts;
    \vspace{-2mm}
    \item {\em  Concatenation}: $\zv_t = [\cv_{1,t}, \cdots, \cv_{M,t}]$. It keeps the all the information of language contexts from multiple teachers, but potentially increases the learning burden in action selection. %
    \vspace{-2mm}
\end{itemize}

\paragraph{Action Selection.}
Actions are predicted based on both the memory vector $\hv_t$ and the aggregated instruction context $\zv_t$.
The action predictor produces a probability for each navigable direction using a bi-linear dot product:
\vspace{-3mm}
\begin{align}
p_k  = \mbox{Softmax}( (\Wmat_z [\hv_t, \zv_t])^{\top}  \Wmat_u \uv_k  )
\label{eq_history}
\end{align}
where $\uv_k$ is the action embedding that represents the $k$-th navigable direction, and $\Wmat_z$ \& $\Wmat_u$ are trainable projection matrices. The action embedding is the concatenation of an visual feature vector (CNN feature vector extracted from the image patch around that view angle or direction) and a 4-dimensional orientation feature vector 
$[\sin \psi ; \cos \psi ; \sin \omega; \cos \omega ]$ where $\psi$ and $\omega$ are the heading
and elevation angles, respectively~\cite{fried2018speaker}.

\begin{figure}[t!]
	\vspace{-0mm}\centering
	\begin{tabular}{c}
		\hspace{-2mm}
		\includegraphics[height=4.3cm]{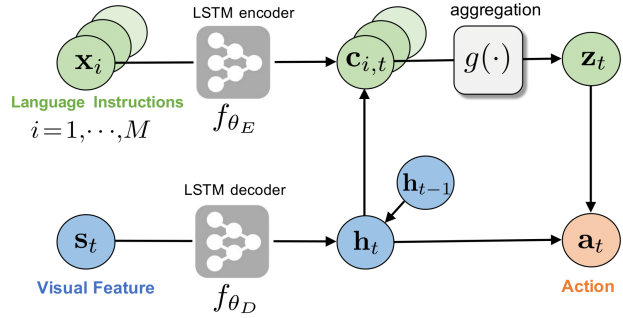}
	\end{tabular}
	\vspace{-2mm}
	\caption{Illustration of learning to navigate with \short{}. 
	The action (\textcolor{red}{red circle}) is selected, based on both the visual scene states (\textcolor{blue}{blue circle}) and textual language instructions (\textcolor{ForestGreen}{green circle}). At time step $t$, \short{} aggregates multiple instructions $\{\xv_i\}_{i=1}^M$ together to generate one action $\av_t$. }
	\vspace{-4mm}
	\label{fig:schemes}
\end{figure}

\paragraph{Learning \& Inference.} 
We summarize the \short{} navigator in Figure~\ref{fig:schemes}. 
At one time step $t$, the agent digests the multiple instructions $\Xcal=\{\xv_i\}_{i=1}^M$ as $\mathcal{C}_t=\{\cv_{i,t}\}_{c=1}^M$, infers a joint representation $\zv_t$, and executes an action $\av_t$. When $M=1$, \short{} simplifies to the single instruction baseline. 
The model parameters $\thetav = \{ \thetav_E, \thetav_D, \Wmat_h, \Wmat_s, \Wmat_z, \Wmat_u\}$ are trained end-to-end with MLE. Note that $\thetav_E$ is shared among different language instructions, and a parameter-free aggregation function $g(\cdot)$ is employed. In testing, this allows to feed an arbitrary number of instruction sequences into the learned policy to produce a single trajectory for evaluation.

%% file: 04_exp.tex
\section{Experiments}
\label{sec:exp}

\paragraph{Dataset.} 
The Room-to-Room (R2R) dataset~\cite{anderson2018vision} is built upon the Matterport3D dataset~\cite{chang2017matterport3d}, which consists of 10,800 panoramic views (each panoromic view has 36 images) and 7,189 trajectories. Each trajectory is paired with three natural language instructions. The R2R dataset consists of four splits: train, validation seen and validation unseen, test unseen. 
At the beginning of each navigation task, the agent starts at a specific location in an environment (one room). The goal of the agent is to follow natural language instructions to navigate to the target location as quickly as possible. 

\paragraph{Implementation.} In all of our experiments, we use a single-layer LSTM for the shared language encoder, and a second single-layer LSTM for the action decoder, with hidden size 512. Following~\cite{fried2018speaker}, we use a panoramic action space. We use the ResNet visual representations provided by~\cite{anderson2018vision}. 
LEO+ is built on top of LEO and trained with the mixture loss (supervised learning and reinforcement learning) and data augmentation. 
We augmented the instructions of the trajectory dataset that \cite{fried2018speaker} provided. Different from speaker-follower, we generate multiple instructions for each trajectory.  
The number of original and augmented instructions for each trajectory is $M = 3$ and $M^{\prime}=3$, respectively. The source code will be made publicly available on GitHub. 

\paragraph{Evaluation Metrics.}
We evaluate our agent on the following metrics:
\begin{itemize}[noitemsep,topsep=2pt] 
\item[\textbf{\texttt{TL}}] \textbf{Trajectory Length} measures the average length of the navigation trajectory.
\item[\textbf{\texttt{NE}}] \textbf{Navigation Error} is the mean of the shortest path distance in meters between the agent's final location and the target location. 
\item[\textbf{\texttt{SR}}] \textbf{Success Rate} is the percentage of the agent's final location that is less than 3 meters away from the target location.
\item[\textbf{\texttt{SPL}}] \textbf{Success weighted by Path Length~\cite{anderson2018evaluation}} trades-off \texttt{SR} against \texttt{TL}. Higher score represents more efficiency in navigation.
\end{itemize}
Among these metrics, \texttt{SPL} is the recommended primary metric, other metrics are considered as auxiliary measures.

\paragraph{Baselines.}
We compare our approach with \emph{nine} recently published systems:
\begin{itemize}[noitemsep,topsep=0pt]
\item \textsc{Random}: an agent that randomly selects a direction and moves five step in that direction ~\cite{anderson2018vision}. 
\item \textsc{S2S-Anderson}: the best performing sequence-to-sequence model using a limited discrete action space, proposed by Anderson {\em et al.} as a baseline for the R2R benchmark~\cite{anderson2018vision}.
\item \textsc{RPA}~\cite{wang2018look}: is an agent which combines model-free and model-based reinforcement learning, using a look-ahead module for planning. 
\item \textsc{Speaker-Follower}~\cite{fried2018speaker}: an agent trained with data augmentation from a speaker model on the panoramic action space.
\item \textsc{Smna}~\cite{ma2019self}: an agent trained with a visual-textual co-grounding module and a progress monitor on the panoramic action space.
\item \textsc{RCM+SIL}~\cite{wang2018reinforced}: an agent trained with cross-modal grounding locally and globally via reinforcement learning.
\item \textsc{Regretful}~\cite{ma2019regretful}: an agent with a trained progress monitor heuristic for search that enables backtracking. 
\item \textsc{Fast}~\cite{ke2019tactical}: an agent which uses a fusion function to combine global and local knowledge to score and compare partial trajectories of different lengths, which enables the agent to efficiently backtrack after a mistake.
\item \textsc{EnvDrop}~\cite{tan2019learning}: an agent is trained with environment dropout, which can generate more environments based on the limited seen environments. 
\item \textsc{PreSS}~\cite{li2019press}: an agent is trained with pre-trained language models and stochastic sampling to generalize well in the unseen environment.

\item \textsc{Prevalent}~\cite{hao2020towards}: a generic agent is pre-trained with image-language-action triples, and fine-tuned on the R2R task.
\end{itemize}


\begin{table}[t!]
\begin{minipage}{.48\linewidth}
    \centering
    \vspace{0mm}
    \hspace{0mm}
    \small
    \begin{tabular}{@{}c@{\hspace{1pt}}c|l@{\hspace{1pt}}l@{}|l@{\hspace{1pt}}l@{}}
    \toprule
    & & \multicolumn{2}{c}{Validation Seen} & \multicolumn{2}{c@{}}{Validation Unseen} \\ 
    & Setting & \texttt{SR} $\uparrow$ & \hspace{-2mm}\texttt{SPL}  $\uparrow$ & \texttt{SR} $\uparrow$ & \hspace{-2mm}\texttt{SPL} $\uparrow$\\ 
    \midrule
    \multirow{2}{*}{(A)}
    & seq2seq\phantom{+} & 51 & 46 & 32 & 25 \\
    & \short\phantom{-} & 52 (\textcolor{blue}{\textbf{+1}}) & 47 (\textcolor{blue}{\textbf{+1}}) & 36 (\textcolor{blue}{\textbf{+4}}) & 31 (\textcolor{blue}{\textbf{+6}})\\
    \hline
    \multirow{2}{*}{(B)}
    & seq2seq\phantom{+} & 49 & 44 & 33 & 26 \\
    & \short\phantom{-} & 63 (\textcolor{blue}{\textbf{+14}}) & 58 (\textcolor{blue}{\textbf{+14}}) & 48 (\textcolor{blue}{\textbf{+15}}) & 41 (\textcolor{blue}{\textbf{+15}}) \\
    \bottomrule
    \end{tabular}
    \caption{Comparison of \short ~and seq2seq. Settings (A) and (B) correspond to single- and multi-instruction respectively, as elaborated in Section Results.}
    \vspace{-3mm}
    \label{tab:greedy_res}
\end{minipage}
\begin{minipage}{.5\linewidth}
\centering
\vspace{0mm}
\hspace{0mm}
\small
\begin{tabular}{@{}c|l@{\hspace{1pt}}l@{}|l@{\hspace{1pt}}l@{}}
\toprule
\multicolumn{1}{c}{} & \multicolumn{2}{c}{Validation Seen} & \multicolumn{2}{c}{Validation Unseen} \\ 
Model &  \texttt{SR} $\uparrow$ & \hspace{-3mm} \texttt{SPL} $\uparrow$ & \texttt{SR} $\uparrow$ & \hspace{-3mm} \texttt{SPL}  $\uparrow$ \\ 
\midrule
SMNA & 63 & 56 & 44 & 30 \\
\phantom{++} + \short & 76 (\textcolor{blue}{\textbf{+13}}) & 68 (\textcolor{blue}{\textbf{+12}}) & 51 (\textcolor{blue}{\textbf{+7}}) & 39 (\textcolor{blue}{\textbf{+9}}) \\
\midrule
EnvDrop (IL) & 48 & 46 & 43 & 40 \\
\phantom{++} + \short & 58 (\textcolor{blue}{\textbf{+10}}) & 56 (\textcolor{blue}{\textbf{+10}}) & 53 (\textcolor{blue}{\textbf{+10}}) & {50} (\textcolor{blue}{\textbf{+10}}) \\
\midrule
EnvDrop (IL+RL) & 55 & 53 & 46 & 43 \\
\phantom{++} + \short & \textbf{64} (\textcolor{blue}{\textbf{+9}}) & \textbf{62} (\textcolor{blue}{\textbf{+9}}) & \textbf{59} (\textcolor{blue}{\textbf{+13}}) & {\textbf{55}} (\textcolor{blue}{\textbf{+12}}) \\
\bottomrule
\end{tabular}
\caption{Improvement on the existing SoTA models (without Data Augmentation).}
\vspace{-3mm}
\label{tab:model_agno}
\end{minipage}
\end{table}


\subsection{Results}
\label{sec:results}
As the key contribution of this work is to develop a new learning paradigm LEO for VLN, we aim to answer
the following research questions via experiments: 
$(\RN{1})$ Is \short{} more effective than the traditional single instruction learning, \emph{even when the total number of provided instructions are controlled to be identical}?
$(\RN{2})$ Is \short{} general enough to allow for easy integration with existing techniques and lead to performance boost?
$(\RN{3})$ How well does \short{} perform when compared with state-of-the-art methods on the unseen test split?

\paragraph{The Effectiveness of LEO.} 
We compare \short{} against the seq2seq greedy agent\footnote{Seq2seq is denoted as baseline agent that has a panoramic action space, \ie the \textsc{Follower} model in the \textsc{Speaker-Follower} (without data augmentation)~\cite{fried2018speaker}.} on the validation splits (seen and unseen). Our agent obtains the aggregated representation of multiple instructions via a parameter-shared encoder and parameter-free pooling function. It allows the model to take an arbitrary number of instructions as input.

We keep the experimental settings the same with seq2seq, except that we add the aggregation function in \short{}.  
Note that there are three different instructions that correspond to one single ground-truth navigation trajectory in the validation/testing dataset splits.
To ensure a fair comparison, we test \short{} and the baseline seq2seq agent in two different evaluation settings:
(A) A single instruction is provided to the agent at a time. Thus, three separate navigation trajectories are generated corresponding to three alternative instructions in this setting. We report the averaged performance over three separate runs.
(B) All three instructions are provided to the agent at once. 
Due to lack of instruction aggregation mechanism in the standard seq2seq models, we report its performance for the single trajectory with maximum likelihood. As for \short{}, the mean-pooling function is used to aggregate the instructions, based on which one single trajectory is generated.
%

The results are summarized in Table~\ref{tab:greedy_res}.
Our approach outperforms seq2seq in both settings. It is interesting to observe that \short{} can provide decent improvement over seq2seq even in setting (A). Even when both agents take one single instruction per navigation, \short{} generalizes better than seq2seq in unseen environments. This demonstrates the advantage of ``multi-view'' learning, where an agent that learns  under the guidance of multiple instructions in training should generalize better. In setting (B), \short{} yields significantly higher performance than seq2seq, in terms of both efficiency (\texttt{SPL}) and success rate. 


\paragraph{LEO as a General Learning Paradigm.}
\short{} can serve as a basic building block for many existing techniques in VLN. We demonstrate this by applying \short{} on two state-of-the-art systems: SMNA~\cite{ma2019self} and EnvDrop~\cite{tan2019learning}. In Table~\ref{tab:model_agno}, with \short{}, both show significant improvement ($50$ and $55$). 

\begin{table*}[t!]
\small
\centering
\begin{tabular}{@{}l@{\hspace{5pt}}c@{\hspace{5pt}}c@{\hspace{5pt}}c@{\hspace{4pt}}cc@{\hspace{6pt}}c@{\hspace{5pt}}c@{\hspace{2pt}}cc@{\hspace{6pt}}c@{\hspace{6pt}}c@{\hspace{3pt}}c@{}}
\toprule
& \multicolumn{4}{c}{Validation Seen} & \multicolumn{4}{c}{Validation Unseen} & \multicolumn{4}{c}{Test Unseen} \\ 
Model & \texttt{TL} $\downarrow$ & \texttt{NE} $\downarrow$ & \texttt{SR} $\uparrow$ & \texttt{SPL} $\uparrow$ & \texttt{TL} $\downarrow$  & \texttt{NE} $\downarrow$ & \texttt{SR} $\uparrow$ & \texttt{SPL} $\uparrow$ & \texttt{TL} $\downarrow$ & \texttt{NE} $\downarrow$ & \texttt{SR} $\uparrow$ & \texttt{SPL} $\uparrow$\\ 
\midrule
\multicolumn{13}{l@{}}{\textit{Approaches that \textbf{\emph{do not}} explore the \emph{test} environments during training, and utilizes \textbf{\emph{one}} instruction during testing}}\\
\textsc{Random}~\cite{anderson2018vision} & \textbf{9.58} & 9.45 & 16 & - & 9.77 & 9.23 & 16 & - & 9.93 & 9.77 & 13 & 12 \\
\textsc{S2S-Anderson}~\cite{anderson2018vision} & 11.33 & 6.01 & 39 & - & \textbf{8.39} & 7.81 & 22 & - & \textbf{8.13} & 7.85 & 20 & 18 \\
\textsc{RPA}~\cite{wang2018look} & - & 5.56 & 43 & - & - & 7.65 & 25 & - & 9.15& 7.53 & 25 & 23 \\
\textsc{Speaker-Follower}~\cite{fried2018speaker} & - & 3.36 & 66 & - & - & 6.62 & 35 & - & 14.82 & 6.62 & 35 & 28\\
\textsc{SMNA}~\cite{ma2019self} & - & - & - & - & - & - & - & - & 18.04 & 5.67 & 48 & 35 \\
\textsc{RCM+SIL}~\cite{wang2018reinforced} & 10.65 & 3.53 & 67 & - & 11.46  & 6.09 & 43 & - & 11.97 & 6.12 & 43 & 38 \\
\textsc{Regretful}~\cite{ma2019regretful} & - & 3.23 & 69 & 63 & - & 5.32 & 50 & 41 & 13.69 & 5.69 & 48 & 40 \\
\textsc{Fast}~\cite{ke2019tactical} & - & - & - & - & 21.17 & 4.97 & 56 & 43 & 22.08 & 5.14 & 54 & 41 \\
\textsc{EnvDrop}~\cite{tan2019learning} & 11.00 & 3.99 & 62 & 59 & 10.70 & 5.22 & 52 & 48 & 11.66 & 5.23 & 51 & 47 \\
\textsc{PreSS}~\cite{li2019press} & 10.57 & 4.39 & 58 & 55 & 10.36 & 5.28 & 49 & 45 & 10.77 & 5.49 & 49 & 45 \\
\textsc{AuxRN(*)}~\cite{zhu2019vision}  & - & 3.33 & 70 & 67 & - & 5.28 & 54 & 50 & - & 5.15 & 55 & 51 \\
\textsc{Prevalent}~\cite{hao2020towards} & 10.32 & 3.67 & 69 & 65 & 10.19 & 4.71 & 58 & 53 & 10.51 & 5.30 & 54 & 51\\
\midrule 
\multicolumn{13}{l@{}}{\textit{Approaches that \textbf{\emph{do not}} explore the \emph{test} environments during training, and utilizes \textbf{\emph{three}} instruction during testing}}\\
\textsc{PreSS}~\cite{li2019press} & 10.35 & 3.09 & 71 & 67 & 10.06 & 4.31 & 59 & 55 & 10.52 & 4.53 & 57 & 53 \\
\textsc{Prevalent}~\cite{hao2020towards} & 10.31 & 3.31 & 67 & 63 & 9.98 & 4.12 & 60 & 57 & 10.21 & 4.52 & 59 & 56 \\
\rowcolor{Gray}
\shortstar{} (Ours) & 10.41 & \textcolor{blue}{\textbf{2.30}} & \textcolor{blue}{\textbf{81}} & \textcolor{blue}{\textbf{78}} & 10.06 & \textcolor{blue}{\textbf{3.35}} & \textcolor{blue}{\textbf{70}} & \textcolor{blue}{\textbf{65}} & 10.24 & \textcolor{blue}{\textbf{3.76}} & \textcolor{blue}{\textbf{65}} & \textcolor{blue}{\textbf{62}} \\
\midrule 
\multicolumn{13}{l@{}}{\textit{Approaches that \textbf{\emph{do}} explore the \emph{test} environments during training, and utilizes \textbf{\emph{one}} instruction during testing}}\\
\textsc{RCM+SIL}~\cite{wang2018reinforced} & 10.13 & 2.78 & 73 & - & 9.12 & 4.17 & 61 & - & 9.48 & 4.22 & 61 & 59 \\
\textsc{EnvDrop}~\cite{tan2019learning} & 9.92 & 4.84 & 55 & 52 & 9.57 & 3.78 & 65 & 61 & 9.79 & 3.97 & 64 & 61 \\
\textsc{AuxRN(*)}~\cite{zhu2019vision} & - & - & - & - & - & - & - & - & - & 3.69 & 68 & 65\\
\midrule
Human & - & - & - & - & - & - & - & - & 11.85 & 1.61 & 86 & 76 \\
\bottomrule
\end{tabular}
\vspace{-1mm}
\caption{Comparison with the previous SoTA methods. \textcolor{blue}{\textbf{Bold}} indicates best value. \shortstar{} with multi-intruction setting achieves near-SoTA scores, which are higher 
all previous SoTA methods that do not explore test enviroments. 
\textbf{(*) indicates unpublished works.}
}
\vspace{-1mm}
\label{tab:main_result}
\end{table*}




%
%

\paragraph{Comparison with SoTA.}
Table~\ref{tab:main_result} compares the performance of our agent against the existing published top systems.\footnote{The full leaderboard is available: https://evalai.cloudcv.org/web/challenges/challenge-page/97/leaderboard/270} With the same data augmentation as the existing work, our \shortstar{} agent significantly outperforms the existing models on nearly all the metrics, includes the two agents which explored test unseen environments (with \textit{one} available instruction) and two systems trained which do not (with access to \textit{three} instructions during testing). 

%% file: 05_analysis.tex

\subsection{Ablation Analysis}
\label{sec:analysis}
We investigate the effects of several alternatives, and perform ablation studies on the base LEO model.



\begin{table}[t!]
\begin{minipage}{.5\linewidth}
    \small
    \centering
    \begin{tabular}{c|rc|rc}
    \toprule
    \multicolumn{1}{c}{} & \multicolumn{2}{c}{Validation Seen} & \multicolumn{2}{c}{Validation Unseen} \\ 
    \texttt{} & \texttt{SR} $\uparrow$ & \texttt{SPL} $\uparrow$ & \texttt{SR} $\uparrow$  & \texttt{SPL} $\uparrow$ \\ 
    \midrule
    Mean & 63 & 58 & \textcolor{blue}{\textbf{48}} & \textcolor{blue}{\textbf{41}} \\
    Max  & 66 & 58 & 41 & 34 \\
    Cat  & 64 & 57 & 36 & 28 \\
    \bottomrule
    \end{tabular}
    \caption{Performance of aggregation schemes.}
    \label{tab:attention}
\end{minipage}
\begin{minipage}{.5\linewidth}
    \centering
    \small
    \vspace{-3mm}
    \begin{tabular}{c|rc|rc}
    \toprule
    \multicolumn{1}{c}{} & \multicolumn{2}{c}{Validation seen} & \multicolumn{2}{c}{Validation Unseen} \\ 
    Encoder &  \texttt{SR} $\uparrow$ & \hspace{-3mm} \texttt{SPL} $\uparrow$ & \texttt{SR} $\uparrow$ & \hspace{-3mm} \texttt{SPL} $\uparrow$ \\ 
    \midrule
    Shared & 63 & 58 & \textcolor{blue}{\textbf{48}} & \textcolor{blue}{\textbf{41}} \\
    Multi-Arm & 62 & 57 & 34 & 28 \\
    \bottomrule
    \end{tabular}
    \caption{Comparison of encoder architectures.}
    \label{tab:parallel_encoders}
\end{minipage}%
\end{table}

\paragraph{Aggregating Schemes.}
%
In Table~\ref{tab:attention} we investigate the impact of various context aggregation schemes detailed in Section \ref{sec:method}. 
A shared encoder is employed for different instructions in this experiment.
The three schemes perform similarly on the seen environments. However, in unseen environments, 
the mean-pooling scheme yields the highest success rate and SPL. This indicates that mean-pooling provides the robust generalization across environments. One might view mean-pooling as a conservative strategy to calculate the representative features over available instructions, compared to the aggressive strategy of max-pooling which only focuses on the most prominent features. Note the concatenation scheme introduces additional trainable parameters, thereby increasing the learning burden for its intermediate layer. Therefore, we consider the mean-pooling aggregation throughout all our experiments by default. 


\paragraph{Multi-Arm Encoders.}
Given a group of instructions-trajectory pairs, our \short{} agent employs a simple strategy to extract the features for different instructions: a shared language encoder.
An alternative is to train a multi-arm encoder, where each arm has its own trainable parameters, and process its own instruction. 
We compare the two strategies in Table~\ref{tab:parallel_encoders}.
Mean-pooling aggregation is used when merging these instruction context. 
The shared encoder is significantly better than the multi-arm encoder on the unseen environments,
though they perform similarly on the seen environments. we conjecture that the parameter-shared encoder does not over-fit on the training dataset as much. Our work focuses on opening the door for the research in joint instruction reasoning for VLN; we leave more advanced choices for encoder design as  future work. 
\begin{figure*}[t!]
\small
$\mathtt{Instruction~{\bf B}}$: {\em Exit the room going straight. \underline{\textcolor{Plum}{Turn right go down the hallway}} until you get to a black bookcase. Turn right \\and continue going down the hallway until you get to a bedroom. Wait at the entrance.} \\
\vspace{-5mm}
\begin{subfigure}[t]{0.9\textwidth}
        \centering
        \vspace{-1mm}
        \includegraphics[scale=0.6]{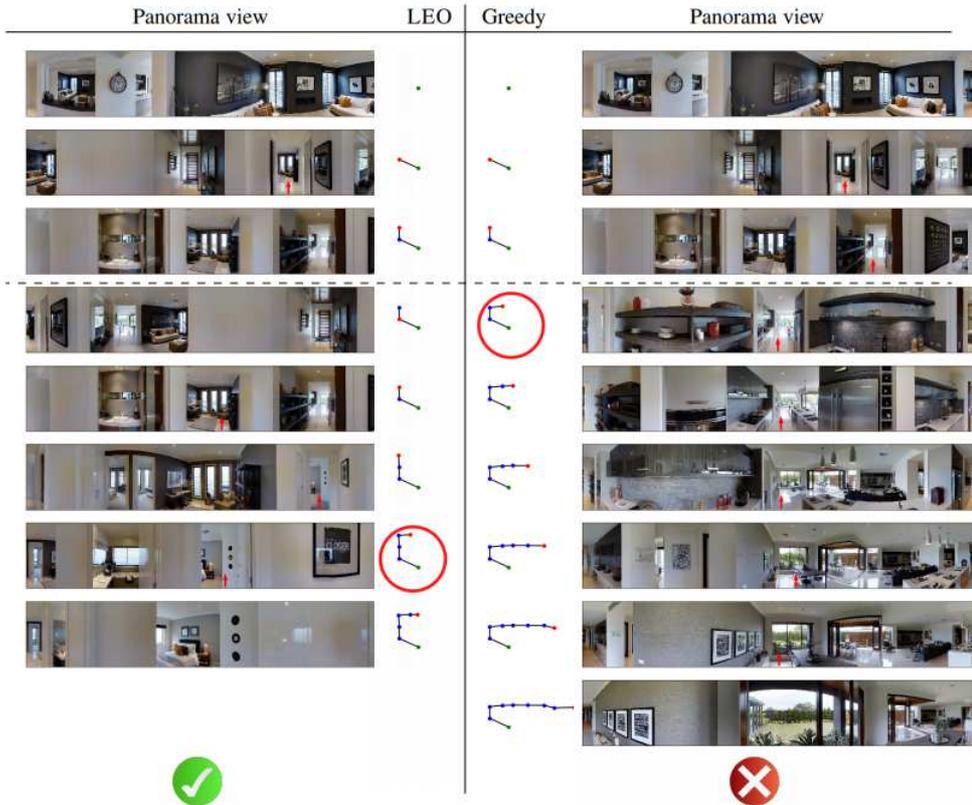}
        \vspace{-2mm}
        \label{fig:}
    \end{subfigure}
    \raggedright
\vspace{3mm}
\caption{Step-by-Step navigation views of \short{}~(left) and \textsc{Speaker-Follower} (right) following the $\mathtt{Instruction~{\bf B}}$ in Figure~\ref{fig:motivation}. For the top three views above the dashed line, actions taken by two agents are identical. 
The views of executing the instruction ``{\em turn right}'' are highlighted for each agent with a red circle: the 7th step of \short{} and the 4th step of \textsc{Speaker-Follower}.}
\label{fig:examples}
\vspace{-2mm}
\end{figure*}

\paragraph{Qualitative Examples.}
We visualize the step-by-step navigation of our \short{} agent and \textsc{Speaker-Follower} in Figure~\ref{fig:examples}, following $\mathtt{Instruction~{\bf B}}$ from Figure~\ref{fig:motivation} in an unseen environment. One may refer the attention heatmap showing the step-by-step agent actions with three instructions in Appendix~\ref{sec:att_vis} for better illustration. 
The two agents take the same actions for the first three steps, as the instruction ``{\em Exit the room going straight}'' is quite specific.  
However, at the fourth step, the \textsc{Speaker-Follower} agent turns right, which eventually leads to a failure. While, our \short{} agent takes the second right at the seventh step and quickly finds the target location. More qualitive examples are provided in Appendix~\ref{sec:more_examples}.


%% file: 02_related.tex
\section{Related Work}
\label{sec:related}

\paragraph{Vision-Language Navigation.} 
%
Most existing approaches to VLN are based on seq2seq architecture~\cite{anderson2018vision}. 
\cite{fried2018speaker} introduced a panoramic action space and a ``speaker" model for data augmentation. 
\cite{ke2019tactical} proposed a novel neural decoding scheme with search to balance global and local information. 
To improve the alignment of the instruction and visual scenes, visual-textual co-grounding attention mechanism  was proposed in~\cite{ma2019self}, which is further improved with a progress monitor~\cite{ma2019regretful}. To improve the generalization of the learned policy to unseen environments, reinforcement learning has been considered, including planning~\cite{wang2018look}, and exploration of unseen environments using a off-policy method~\cite{wang2018reinforced}. \cite{tan2019learning} proposed an environment dropout to generate more environments based on the limited environments, so that it can generalize well to unseen environments. All these approaches assume a training regime where a single instruction is considered in isolation from other related instructions. The proposed \short{} paradigm presents the first work to leverage the joint representation of multiple related instructions in one navigation, and can be easily combined with previous methods to achieve improved performance.


\paragraph{Multi-View Learning.} 
Multi-view learning has been applied successfully in a number of real-world applications~\cite{xu2013survey}, such as web-page classification~\cite{blum1998combining}, information retrieval~\cite{wang2010multi}, and face detection~\cite{li2002statistical}. It has recently been integrated with deep neural networks for flexible representation learning~\cite{wang2015deep,kan2016multi}.
By exploring the consistency and complementary properties of different views, \short{}  leads to more effective generalization compared to single-view learning~\cite{sridharan2008information}. The success of VLN requires precise reasoning over highly variable and under-specified language instructions, in order to ground them to the visual environment and action prediction. \short{} demonstrates the advantage of multi-view learning, and exhibits strong generalization in unseen environments.

%% file: 07_conclusion.tex
\section{Conclusion}
\label{sec:con}
We present \short{}, a new training paradigm that leverages mutual agreement across instruction variants. This allows for more effective use of the limited training data to improve generalization to the previously unseen environments. Empirical results on the R2R benchmark demonstrate that \short{} significantly improves over the traditional single instruction paradigm. Further, \short{} can be easily plugged into many existing models to boost their performance, as we demonstrate via \shortstar{}, a navigator enhanced with data augmentation/ It improves the agent's generalization ability in unseen environments when a single instruction is used in testing, and can largely boost the performance when multiple instructions are avaiable.
%



%% file: 08_supp.tex
\section{Appendices}

\subsection{Proof of entropy reduction}\label{sec:proof_of_entr}
\label{sec_supp:entropy}
\begin{align}
& H(\tauv | \Xcal)  \\
& = - \int_{\tauv, \Xcal}
p(\tauv, \Xcal ) \log p(\tauv | \xv_1, \cdots, \xv_M )  
\\
& 
=- \int_{\tauv, \Xcal}
p(\tauv, \Xcal ) 
\log  \left[ \frac{p(\tauv, \xv_1, \cdots, \xv_M )p(\xv_1) }{p(\xv_1, \cdots, \xv_M ) p(\tauv, \xv_1 ) }
 \right]  \nonumber \\
& -\int_{\tauv, \Xcal}
p(\tauv, \Xcal ) 
\log \left[ \frac{ p(\tauv, \xv_1  ) }{p(\xv_1 )   }
 \right] \\
&= - I (\tauv, \xv_2, \cdots, \xv_M  | \xv_1 ) + H(\tauv | \xv_1)\\
&\le H(\tauv | \xv_1) 
\end{align}

\subsection{Attention Visualization}\label{sec:att_vis}
\begin{figure*}[h]
\centering
\includegraphics[scale=0.5]{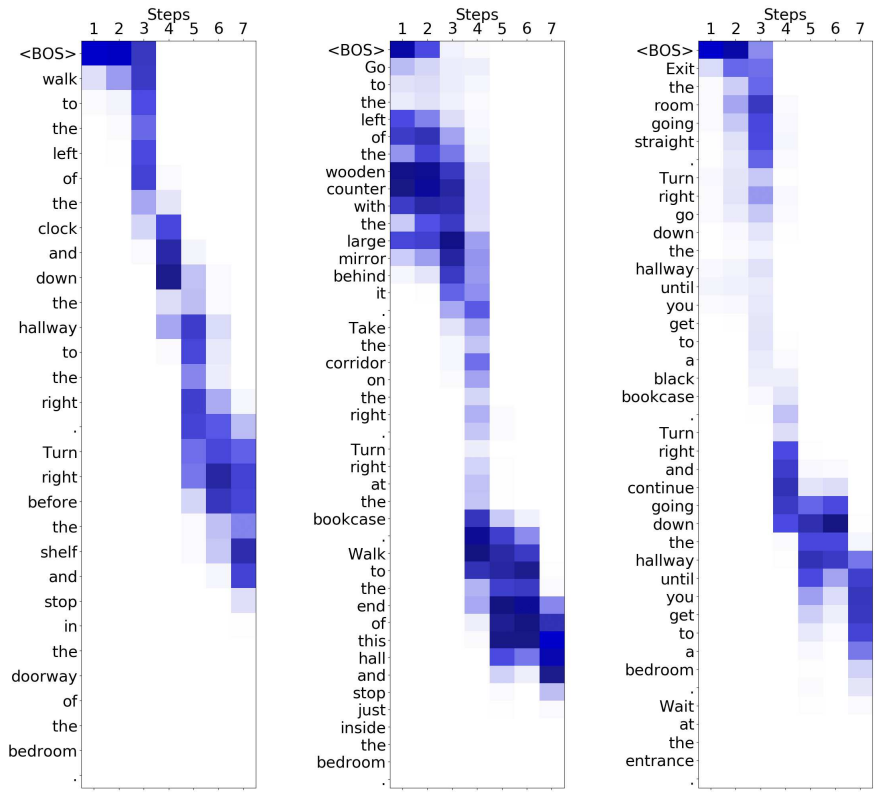}
\caption{An attention heatmap showing the step-by-step agent actions with three instructions.}
\label{fig:attention_hotmap}
\end{figure*}

We visualize three attention heatmaps between the language instruction and action trajectory in Figure~\ref{fig:attention_hotmap}.
Note that these three instructions correspond to one single trajectory, with the step-by-step top-down view in Figure~\ref{fig:motivation} and the panorama view in Figure~\ref{fig:examples}. 
The three instructions provide complementary information to guide the agent to navigate towards the target location.
For example, at the first two steps, the agent attends to a wide range of tokens in instruction B (many elements in the first two columns are highlighted), meaning the agent has difficulties in understanding which words of instruction B to focus on. However, when taking the same action, the agent can clearly concentrate on a few words (such as ``walk'' and ``exit'') at the beginning of instruction A and C.

\subsection{Examples of Successes and Failures}\label{sec:more_examples}
\begin{figure*}[t]
\centering
\begin{tabular}{p{6cm}p{6cm}p{6cm}}
\multicolumn{3}{p{16.0cm}}{ 
$\mathtt{Instruction~{\bf 1}}$: \textit{Exit the bathroom, and walk through the closet. Make a left just before the bed. Exit the bedroom, and make a right. Walk through the open bedroom door on the right. Wait in the door's threshold.}
}\\
\multicolumn{3}{p{16.0cm}}{ 
\textcolor{Plum}{\bf
$\mathtt{Instruction~{\bf 2}}$: \textit{Leave the bathroom and closet. Exit the bedroom and take a right. Enter the room on the right and stop in the doorway.}}
}\\
\multicolumn{3}{p{16.0cm}}{ 
$\mathtt{Instruction~{\bf 3}}$: \textit{Walk straight ahead until you reach the bed. Turn left and walk out the door to the left. Once out, turn right and then enter the door on your right and stop once you enter.}
}
\end{tabular}

\includegraphics[scale=0.75]{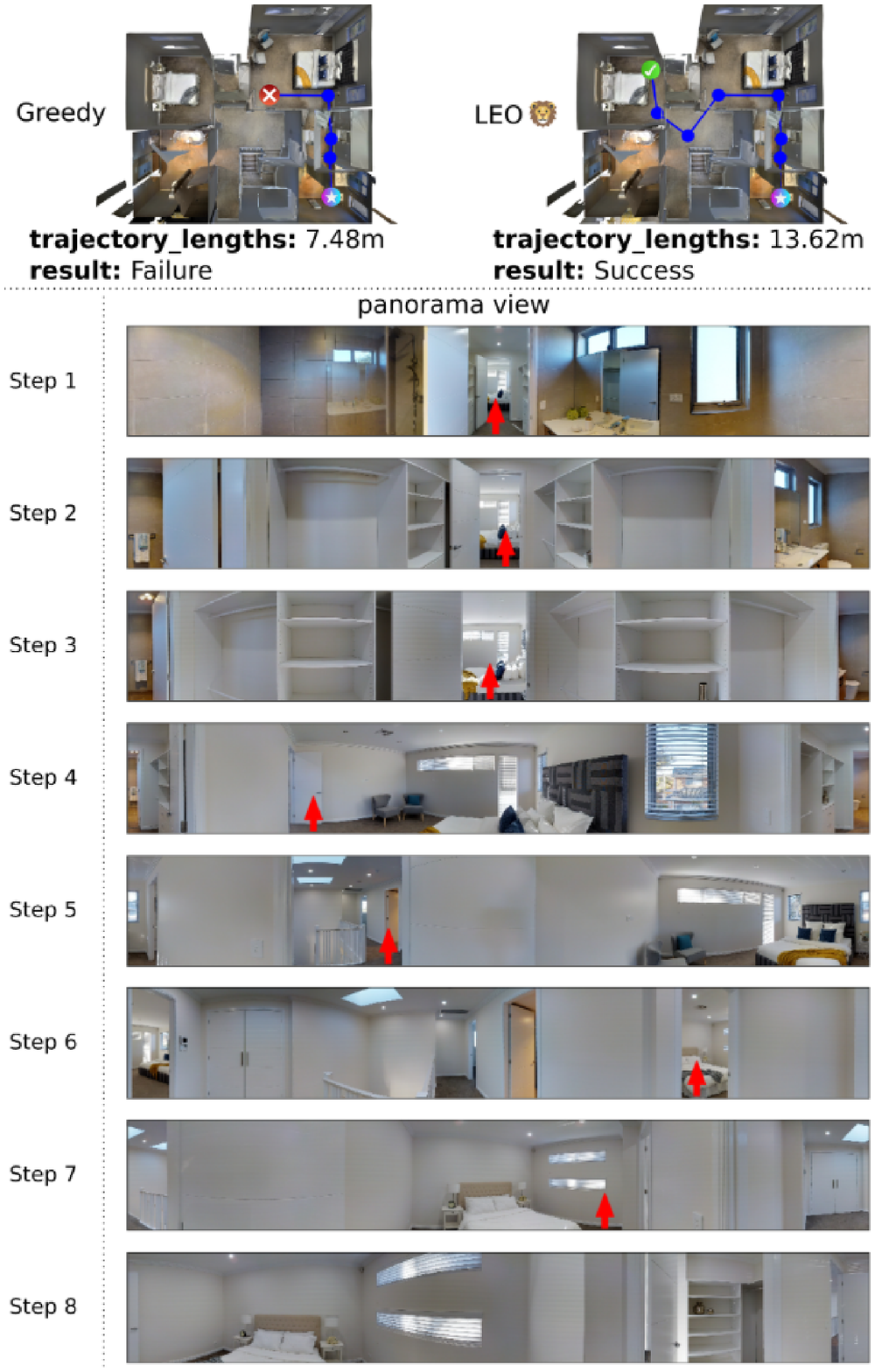}
\caption{Top-Down view and Step-by-Step navigation view of Example (path 2489, instruction 2) in an unseen environment of R2R. \protect\inlinegraphics{img/start-c.eps} indicates the Start, \protect\inlinegraphics{img/stop-yes-c.eps} indicates the Target, \protect\inlinegraphics{img/stop-no-c.eps} indicates the incorrect end point. \textcolor{red}{Red} arrow indicates the direction to go next.}
\label{fig:success_1}
\end{figure*}

\begin{figure*}[t]
\centering

\begin{tabular}{p{6cm}p{6cm}p{6cm}}
\multicolumn{3}{p{16.0cm}}{ 
\textcolor{Plum}{\bf
$\mathtt{Instruction~{\bf 1}}$: \textit{Exit the bedroom, enter the bathroom, wait at the toilet.}
}}\\
\multicolumn{3}{p{16.0cm}}{ 
$\mathtt{Instruction~{\bf 2}}$: \textit{Walk out of the bedroom and take a right into the bathroom. In the bathroom take your first right into the water closet and stop in front of the door.}
}\\
\multicolumn{3}{p{16.0cm}}{ 
$\mathtt{Instruction~{\bf 3}}$: \textit{Walk through the archway to the right and into the bathroom. Wait in the room to the right with the toilet.}}
\end{tabular}

\includegraphics[scale=0.80]{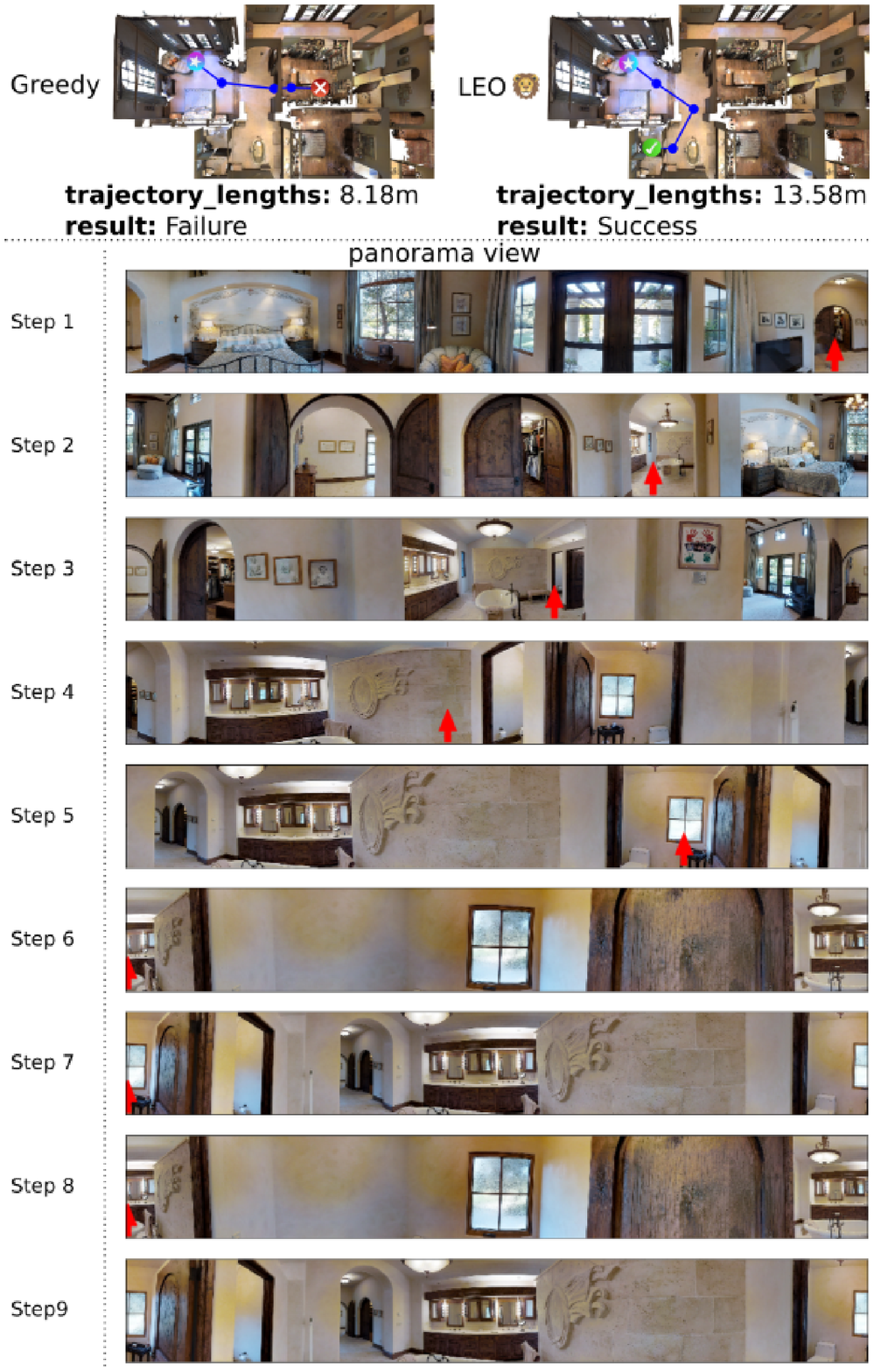}
\caption{Top-Down view and Step-by-Step navigation view of Example (path 321, instruction 1) in an unseen environment of R2R. \protect\inlinegraphics{img/start-c.eps} indicates the Start, \protect\inlinegraphics{img/stop-yes-c.eps} indicates the Target, \protect\inlinegraphics{img/stop-no-c.eps} indicates the incorrect end point. \textcolor{red}{Red} arrow indicates the direction to go next.}
\label{fig:success_2}
\end{figure*}

\begin{figure*}[t]
\centering

\begin{tabular}{p{6cm}p{6cm}p{6cm}}
\multicolumn{3}{p{16.0cm}}{ 
$\mathtt{Instruction~{\bf 1}}$: \textit{Go straight into the door in front of you, turn left and then turn left again to go into the bathroom. Wait by the second sink.}
}\\
\multicolumn{3}{p{16.0cm}}{
\textcolor{Plum}{\bf
$\mathtt{Instruction~{\bf 2}}$: \textit{Walk down the hallway past the painting and along the mirrored wall into the bedroom. Walk beside the bed and into the bathroom. Wait inside the bathroom, next to the shower.}}
}\\
\multicolumn{3}{p{16.0cm}}{ 
$\mathtt{Instruction~{\bf 3}}$: \textit{Head into the bedroom. Turn left and go into the bathroom. Stop in front of the shower.}
}
\end{tabular}

\includegraphics[scale=0.80]{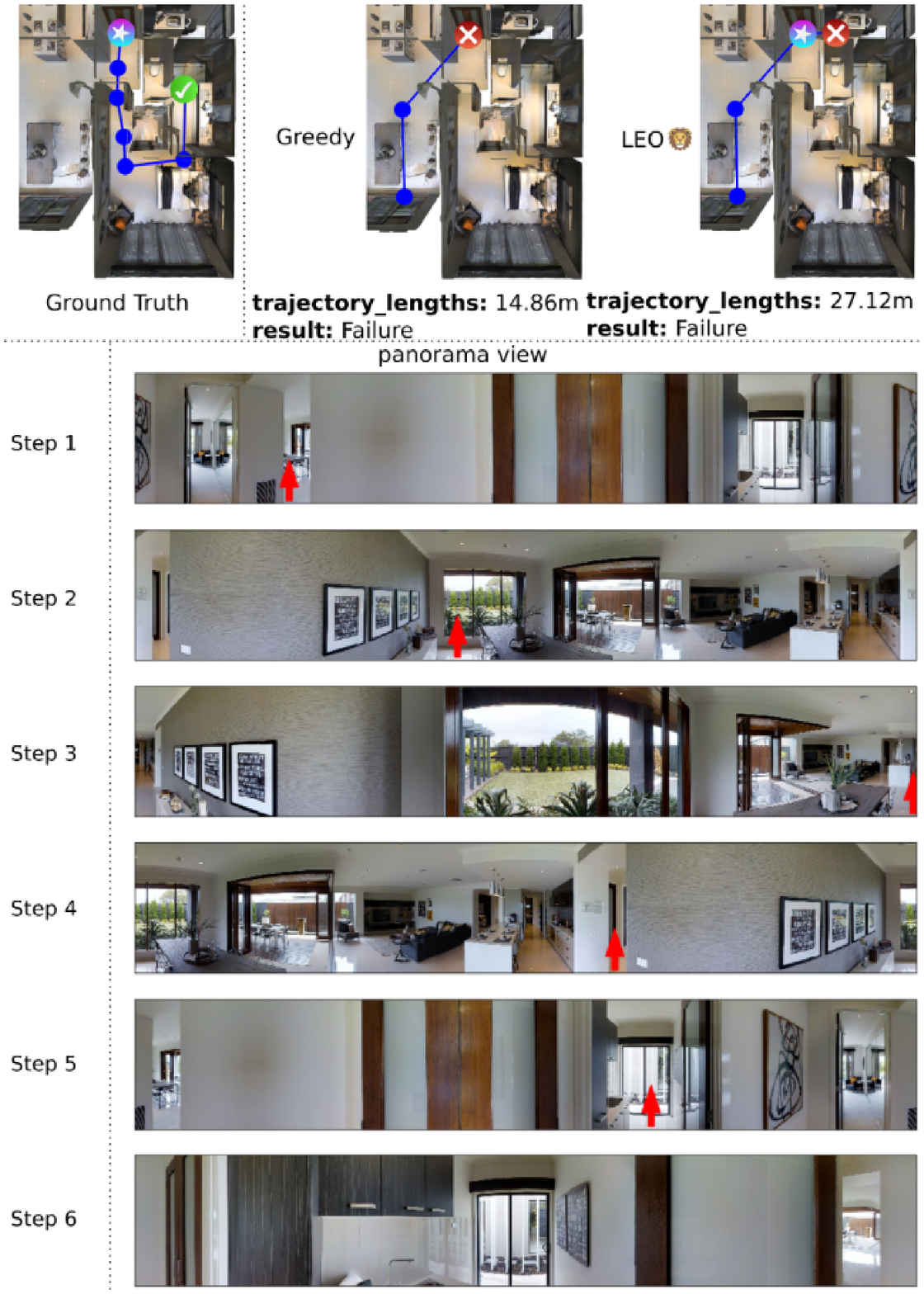}
\caption{Top-Down view and Step-by-Step navigation view of Example (path 2755, instruction 2) in an unseen environment of R2R. \protect\inlinegraphics{img/start-c.eps} indicates the Start, \protect\inlinegraphics{img/stop-yes-c.eps} indicates the Target, \protect\inlinegraphics{img/stop-no-c.eps} indicates the incorrect end point. \textcolor{red}{Red} arrow indicates the direction to go next.}
\label{fig:failure_1}
\end{figure*}

\begin{figure*}[t]
\centering

\begin{tabular}{p{6cm}p{6cm}p{6cm}}
\multicolumn{3}{p{16.0cm}}{
\textcolor{Plum}{\bf
$\mathtt{Instruction~{\bf 1}}$: \textit{Walk out into the yard. Take a left. Take another left, and go around the house. Stop before you go up the second stair.}}
}\\
\multicolumn{3}{p{16.0cm}}{ 
$\mathtt{Instruction~{\bf 2}}$: \textit{Go past the table and chairs and down the steps. Turn left and then turn left to go around the house. Wait there.}
}\\
\multicolumn{3}{p{16.0cm}}{ 
$\mathtt{Instruction~{\bf 3}}$: \textit{Walk off the deck, turn left to walk around the house. Stop and wait near the window.}
}
\end{tabular}

\includegraphics[scale=0.8]{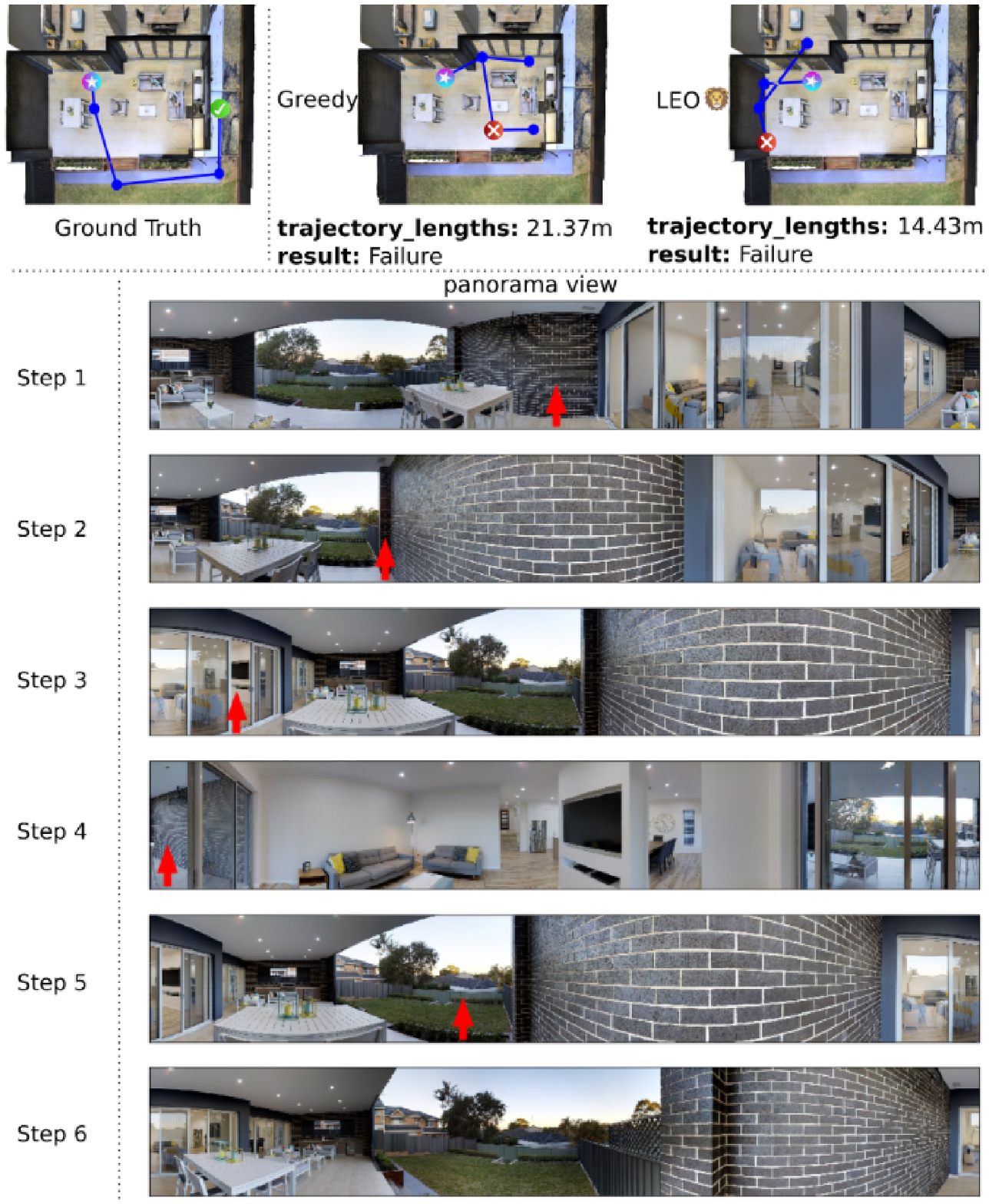}
\caption{Top-Down view and Step-by-Step navigation view of Example (path 5476, instruction 1) in an unseen environment of R2R. \protect\inlinegraphics{img/start-c.eps} indicates the Start, \protect\inlinegraphics{img/stop-yes-c.eps} indicates the Target, \protect\inlinegraphics{img/stop-no-c.eps} indicates the incorrect end point. \textcolor{red}{Red} arrow indicates the direction to go next.}
\label{fig:failure_2}
\end{figure*}

We provide examples of successful trajectories ( in Figures~\ref{fig:success_1} and \ref{fig:success_2}) and failures (in Figure \ref{fig:failure_1} and \ref{fig:failure_2}). The highlighted instruction is executed by the baseline agent.

For each success case, the top-down views highlight the trajectories of Greedy vs. \short, where Greedy agent fails, \short{} succeeds. And we also provide a step-by-step navigation view of \short.

For each failure case, we show the top-down views for the trajectories of Ground-Truth, Greedy  and \short{}, respectively.   We also provide a step-by-step navigation view for \short. Interestingly, on the cases that \short{} agent fails, the greedy agent would usually fail, too.


